\documentclass[manuscript,screen]{acmart}

\usepackage{amsmath,amssymb,amsfonts}

\usepackage{algorithmic}
\usepackage{graphicx}
\usepackage{textcomp}

\usepackage{wrapfig}
\usepackage{times}
\usepackage{epsfig}

\usepackage{gensymb}

\usepackage{multirow}
\usepackage{makecell}
\usepackage{verbatim}

\usepackage{threeparttable}
\usepackage{colortbl}

\AtBeginDocument{%
  \providecommand\BibTeX{{%
    \normalfont B\kern-0.5em{\scshape i\kern-0.25em b}\kern-0.8em\TeX}}}

\setcopyright{acmcopyright}
\copyrightyear{2021}
\acmYear{2021}
\acmDOI{ }

\begin{document}

\title{A Benchmark for Gait Recognition under Occlusion Collected by Multi-Kinect SDAS}

\author{Na Li}
\email{cvlina@mail.nwpu.edu.cn}
\orcid{0000-0002-6460-719X}

\author{Xinbo Zhao}
\orcid{0000-0003-2618-9289}
\email{xbozhao@nwpu.edu.cn (Corresponding author)}
\affiliation{
  \institution{ National Engineering Laboratory for Integrated Aero-Space-Ground-Ocean Big Data Application Technology, School of Computer Science and Engineering, Northwestern Polytechnical University}
    \city{Xi’an}
    \country{China}
  \postcode{710072}
}

\renewcommand{\shortauthors}{Na Li and Xinbo Zhao}


\begin{abstract}
Human gait is one of important biometric characteristics for human identification at a distance. In practice, occlusion usually occurs and seriously affects accuracy of gait recognition. However, there is no available database to support in-depth research of this problem, and state-of-arts gait recognition methods have not paid enough attention to it, thus this paper focuses on gait recognition under occlusion. We collect a new gait recognition database called OG RGB+D database, which breaks through the limitation of other gait databases and includes multimodal gait data of various occlusions (self-occlusion, active occlusion, and passive occlusion) by our multiple synchronous Azure Kinect DK sensors data acquisition system (multi-Kinect SDAS) that can be also applied in security situations. Because Azure Kinect DK can simultaneously collect multimodal data to support different types of gait recognition algorithms, especially enables us to effectively obtain camera-centric multi-person 3D poses, and multi-view is better to deal with occlusion than single-view. In particular, the OG RGB+D database provides accurate silhouettes and the optimized human 3D joints data (OJ) by fusing data collected by multi-Kinects which are more accurate in human pose representation under occlusion. We also use the OJ data to train an advanced 3D multi-person pose estimation model to improve its accuracy of pose estimation under occlusion for universality. Besides, as human pose is less sensitive to occlusion than human appearance, we propose a novel gait recognition method SkeletonGait based on human dual skeleton model using a framework of siamese spatio-temporal graph convolutional networks (siamese ST-GCN). The evaluation results demonstrate that SkeletonGait has competitive performance compared with state-of-art gait recognition methods on OG RGB+D database and popular CAISA-B database. This database will be available at \url{https://github.com/cvNXE/OG-RGB-D-gait-database}.
\end{abstract}



\maketitle

\section{Introduction}
Gait recognition contributes to discriminate individuals by their walking way. In contrast to other biometric features, gait has the advantage of being recognized from a long distance away. Therefore, there are bright prospects for gait recognition in numerous applications such as visual surveillance, forensic identification, and national security. Unfortunately, occlusion usually occurs in gait recognition in practice (e.g. long clothing or large belongings could easily cover human body, two or more person walking in a group, or self-occlusion) that seriously affects the extraction of gait features. Occlusion is a crucial challenge for gait recognition, but state-of-arts gait recognition methods \cite{chao2019gaitset,fan2020gaitpart,lin2020gait} have not paid enough attention to it and there is no available database for further research on this problem. Therefore, this paper focuses on gait recognition under occlusion.

\begin{figure}[!h]
\centering

\includegraphics[width=0.85\linewidth]{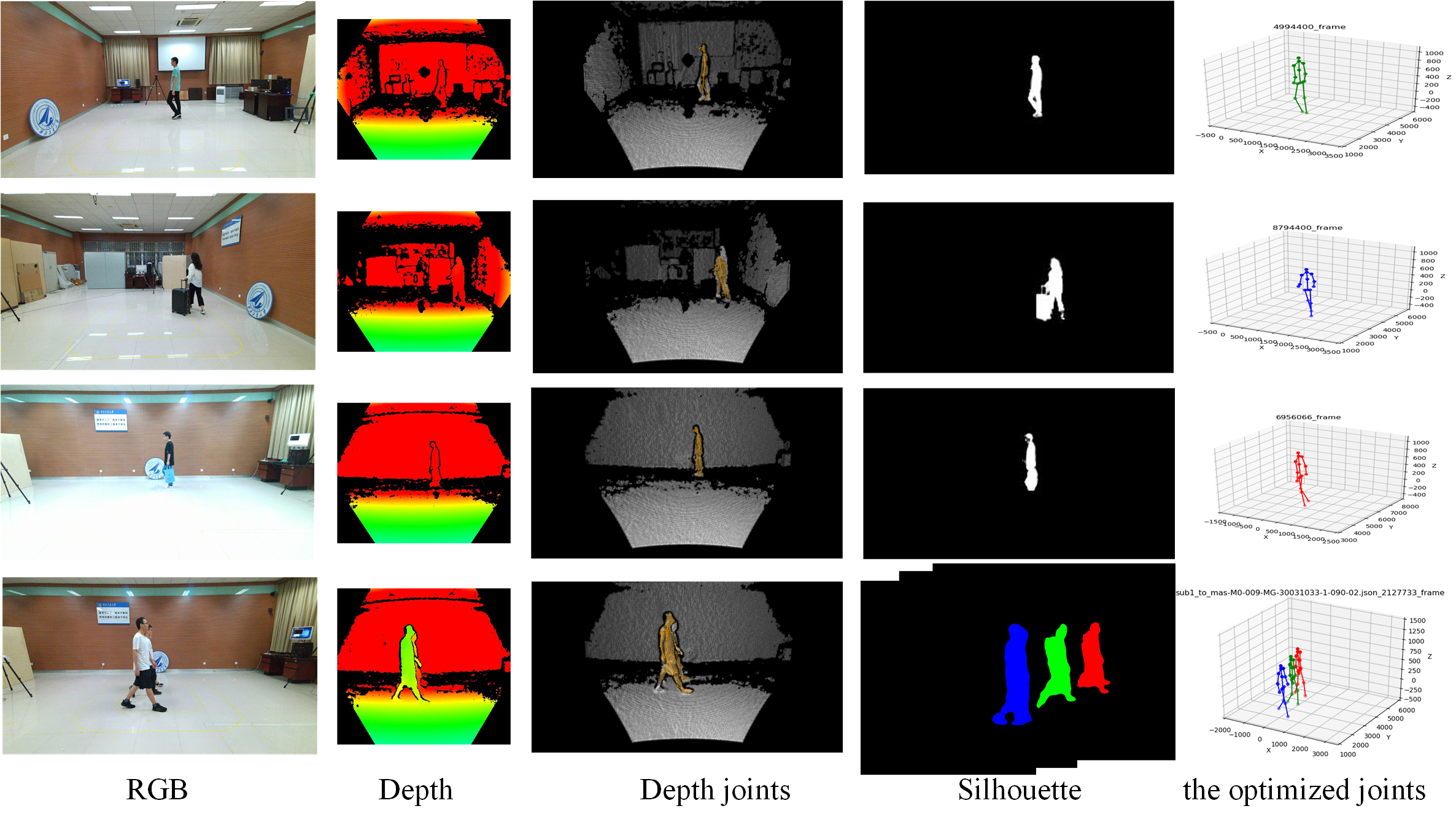}
\centering
\caption{Data modalities provided by the OG RGB+D database.} 

\label{fig1}
\end{figure}
Firstly, for obtaining accurate human pose representation unaffected occlusion and simultaneously collect multimodal data to support different types of gait recognition methods, we design a data acquisition system multi-Kinect SDAS composed of hardware system, acquisition software, and a multi-view data fusion processing method. Although significant improvement \cite{rogez2017lcr,2020VIBE,mehta2017monocular,mehta2018single} has been achieved recently in multi-person 3D pose estimation,  most of them only estimate the relative 3D pose to a reference point in the body, which results in losing the location of each person in the scene and incorrect pose estimation. Advanced pose estimation method \cite{moon2019camera}is proposed to solve this problem. However, when the occlusion is serious in daily life, such as when a group of people walking together, it is still a challenge to estimate the accurate human pose. Therefore, we design the multi-Kinect SDAS to obtain real-time camera-centric multi-person 3D poses that are not affected by occlusion, which helps to extract effective gait features, and this system can also be applied to all occasions where the human pose needs to be accurately estimated. Besides, these accurate pose data can be used as groundtruth to help pose estimation methods improve their performance under occlusion that is beneficial to the popularization of gait recognition. We use the OJ data of OG RGB+D database to train the pose estimation model\cite{moon2019camera}, and experiments show that it can effectively improve its accuracy of camera-centric multi-person 3D poses estimation.

Secondly, we collect OG RGB+D database under various occlusions in daily life by our multi-Kinect SDAS to break through the limitation of other gait databases, including self-occlusion (eight views), active occlusion (backpack, carried small belongings, carried large belongings, heavy coat, long coat or long skirts), and passive occlusion (three subjects walking together). In this way, OG RGB+D database does not simply consider a variety of occlusion covariates, and unlike other databases which only contain single-gait data, it also contains multi-gait data. More importantly, there are multimodal gait data in  OG RGB+D database(as shown in Fig.\ref{fig1}), which can support the research of different types of gait recognition algorithms. In addition, we also collect the gait data when people turn, which is ignored by other gait databases.

Finally, we propose a novel model-based gait recognition method SkeletonGait using siamese ST-GCN, which learns more discriminative gait information from human dual skeleton model composed of 3D joints and anthropometric features. The experimental results demonstrate that SkeletonGait greatly outperforms advanced model-based methods and performs superior to state-of-art appearance-based methods when occlusions seriously affect people’s appearance.
The contributions of this work are the following:

(1) We design a data acquisition system multi-Kinect SDAS to obtain accurate human pose representation unaffected occlusion and simultaneously collect multimodal gait data to support different types of gait recognition methods.

(2) A benchmark for gait recognition under occlusion, OG RGB+D database, is introduced in this paper, considering a variety of occlusion covariates, especially including multi-gait data and turning data which are not available in other databases.

(3) We propose a novel model-based gait recognition method SkeletonGait which learns more discriminative gait information from human dual skeleton model, and receives satisfactory performance compared with state-of-art gait recognition methods.

\section{Related work}

Here, we briefly review Azure Kinect DK sensor, available gait databases, state-of-art gait recognition methods, multi-person 3D pose estimation in this section.

\subsection{Azure Kinect DK}
By integrating an RGB and a depth camera, Microsoft Kinect sensor enables us to obtain effective 3D structure information of the objects and scenes, which has been identified as a portable and cost-effective device for gait assessment \cite{Wang2021Gait,Movement2018}. Azure Kinect DK (Developer Kit), as the latest version of Kinect, was launched in 2019. Like Kinect v2, it also utilizes the ToF principle to estimate depth information, but offers significantly higher accuracy than other commercially available cameras \cite{brown2020automated}. Compared with previous versions, it has a new improved body tracking SDK based on deep learning technologies, which allows for tracking the 3D positions and orientations of 32 human joints. What's more, it is updated to include synchronization ports which can be used to synchronize multiple Azure Kinect DK sensors and is helpful to deal with occlusion.

\subsection{Gait recognition databases}

After the release of the Kinect v1 and Kinect v2, several gait databases \cite{andersson2015person,baisware2019review,borras2012depth,sivapalan2011gait,hofmann2014tum,wang2016gait} have been collected to support model-based gait recognition. They consider various factors such as speed\cite{sivapalan2011gait,baisware2019review}, carrying\cite{sivapalan2011gait,baisware2019review,hofmann2014tum,wang2016gait}, view\cite{borras2012depth,hofmann2014tum} except occlusion caused by clothing and multi-gait, which are very common in practice. 

Recently, focus of research on gait recognition is based on databases collected by RGB cameras because they contain more covariates. CASIA-B database\cite{yu2006framework} is the most adopted database because it has 11 view variations along with clothing and carrying covariate variations, but the clothing and belongings in CASIA-B cover the body slightly. There are a series of gait database \cite{makihara2012isir} separately considering different covariates conditions (such as view, clothing, speed transition and carrying) collected by Osaka University (OU). But they have same limitations as CASIA-B, and because of privacy issues, they only publish silhouettes of gait, which can well support appearance-based algorithms but are not available for model-based algorithms. Currently, OU presents OUMVLP-Pose \cite{An_TBIOM_OUMVLP_Pose} for model-based gait recognition methods, which only include view variations. In addition, these popular databases mentioned above only collect single-gait, ignoring the passive occlusion of multi-gait. TUM-IITKGP\cite{hofmann2011gait} focuses on the problem of static inter object occlusion (hands in pocket, backpack and gown) and dynamic occlusion (two people walking past) with 90\degree view. However, these conditions actually have limited occlusion to human gait.

Therefore, we collect OG RGB+D database including multi-modal data of various occlusion factors to both support model-based gait recognition methods and appearance-based ones. Our database also includes gait data of people turning when walking, which are not available in other databases. Moreover, our data collection environment is closer to daily life, that is, the background is more complex, there are people's shadows and reflections in the floor, and light changes are introduced.

\subsection{Gait recognition }

Gait recognition methods can be broadly classified into two categories: model-based ones and appearance-based ones, respectively taking body models and silhouettes as the gait features.

Using the body motion to recognize gait is straightforward and reasonable, however, accurately locating and tracking human body parts is a tough challenge model-based methods. Due to the emergence of depth cameras such as Kinect, the skeleton-based gait recognition method has gradually increased\cite{andersson2015person,yang2016relative,sun2018view}, but many of them are identified on the self-built database, the generalization performance is not good enough. \cite{PTSN,PTSN-3D,liao2020model} focus on extraction of high-dimensional gait features from the manual skeleton features for recognition. JointsGait\cite{Li2020} takes raw human 2D joints as the input of GCN, which verifies the feasibility of gait recognition using joints. In this paper, we focus on learning more discriminative gait information from human 3D joints.

Recently, GaitSet \cite{chao2019gaitset} regards gait as an independent silhouette set rather than continuous silhouettes \cite{wu2016comprehensive}, and surpasses previous state-of-the-arts\cite{wu2016comprehensive,wu2015learning,SPAE,GaitGAN,MGAN,Gaitganv2}. GaitPart\cite{fan2020gaitpart} enhances the extraction of local information based on GaitSet\cite{chao2019gaitset}. MT3D\cite{lin2020gait} exploits multiple 3D CNNs to integrate temporal information and further improves the accuracy of gait recognition.

\subsection{ Multi-person 3D pose estimation }

Multi-person 3D pose estimation has achieved great progress in recent years. LCR-Net \cite{rogez2017lcr} estimates multi-person 3D pose by first locating people, then classifying the detected human into anchor-poses, and finally refining the anchor-poses. Mehta et al. \cite{mehta2017monocular} introduce a pose-map formulation to support multi-person pose inference. VIBE\cite{2020VIBE}is proposed to estimate multi-person 3D pose and shape through an adversarial learning framework. However, none of these method estimate camera-centric 3D human poses which results in losing important location information. Moon et al.\cite{moon2019camera}firstly propose a learning-based, camera distance-aware method for 3D multi-person pose estimation composed of RootNet and root-relative PoseNet, which significantly outperforms previous methods on publicly available datasets.

\section{ Multi-Kinect SDAS }

The working distance of one Kinect DK is limited, and joints estimation may be incredible due to occlusion. Therefore, we design the multi-Kinect SDAS to collect synchronous multi-view gait data to deal with this problem, which is consists of hardware system, acquisition software and multi-view data fusion processing method. 

\begin{figure*}[htb]
\begin{center}
\includegraphics[width=1\linewidth]{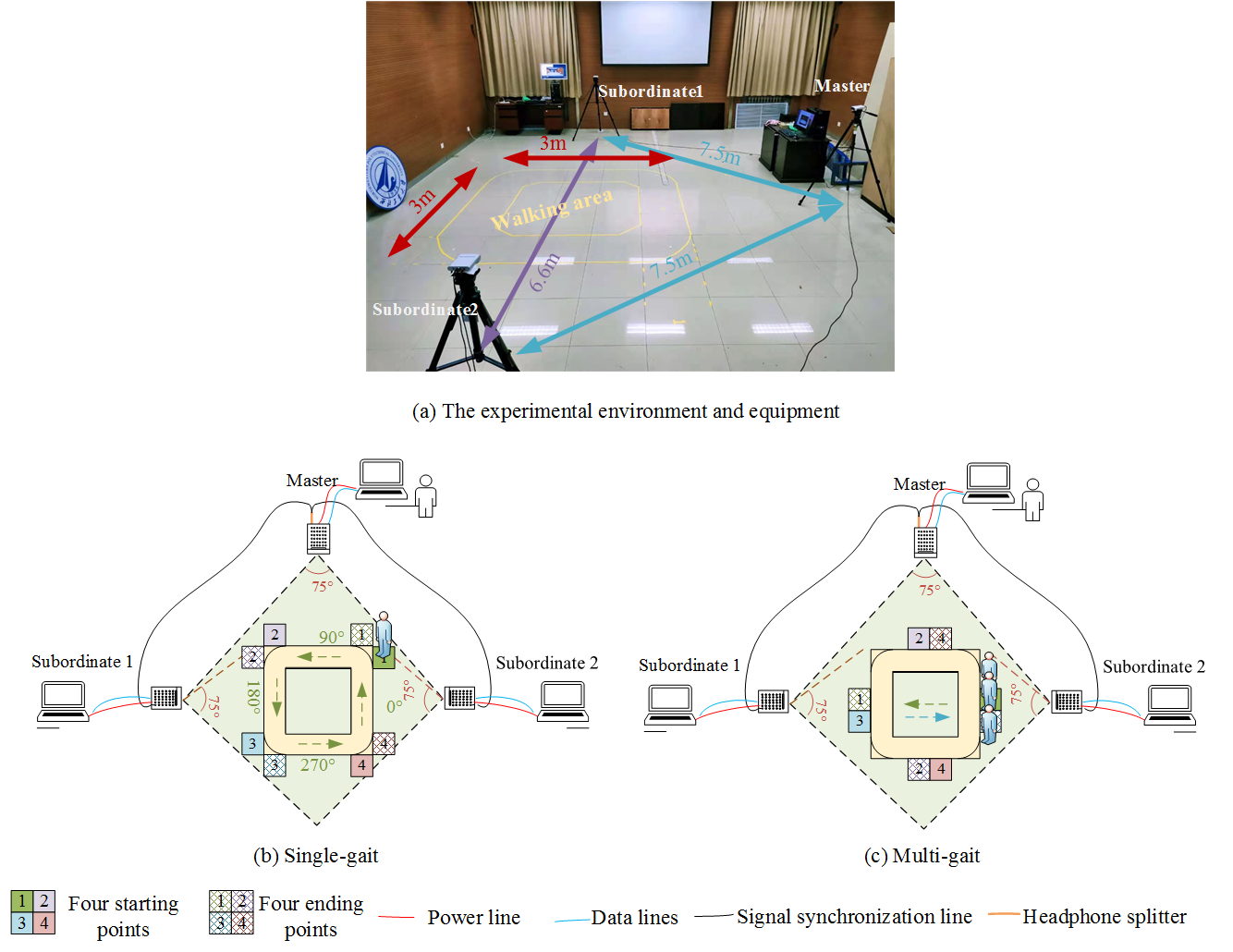}
\end{center}
\caption{Data acquisition processing of the OG RGB+D database by multi-Kinect SDAS. (a) The experimental environment and equipment. The hardware system of multi-Kinect SDAS is mainly composed of three synchronous latest Azure Kinect DK sensors, one is the master device, the other two are subordinate devices. They are respectively supported by a computer with 8G 1080ti GPU and connected by star configuration. (b) When collecting single-gait, the subject walks counter clockwise along the square trajectory. (c) Subjects walk straight when collecting multi-gait. The square filled with solid color is the starting point, and the grid filled square is the corresponding end point.}
\label{fig2}
\end{figure*}

\textbf{Hardware system} The hardware system of multi-Kinect SDAS is mainly composed of three synchronous latest Azure Kinect DK sensors, one is master device, the other two is subordinate devices. As shown in Fig. \ref{fig2}, to achieve accurate synchronous acquisition, the master device and two subordinate devices are placed in an isosceles triangle, respectively supported by a computer with 8G 1080ti GPU and connected by star configuration: the synchronous acquisition signal is extracted from the 3.5-mm Sync out synchronization port of the master sensor by a headphone splitter, then is transmitted to the Sync synchronization ports of two subordinate devices by two 3.5mm audio cables. They are all placed at 1.5 m height and are facing downwards at an angle of rough 10\degree.

The RGB Camera Resolution (H$\times$V) of each device adopts 1280$\times$720, the storage format is MJPEG, and the corresponding Nominal FOV (Field-of-view) is 90\degree$\times$59\degree (H$\times$V). And depth camera adopts NFOV ( Narrow field-of-view depth mode) 2$\times$2 binned (SW) mode, whose FoI (Field of Interest) is 75\degree$\times$65\degree, operating range is 0.5 m-5.46 m. Due to the horizontal FOV of RGB Camera is bigger than the horizontal FOV of the depth camera, so the green background in Fig. \ref{fig2} is a schematic diagram of the operating areas composed of three depth cameras of Kinect DK devices, which completely cover the subject's walking area.

\begin{figure}[htb]
\begin{center}

\includegraphics[width=0.5\linewidth]{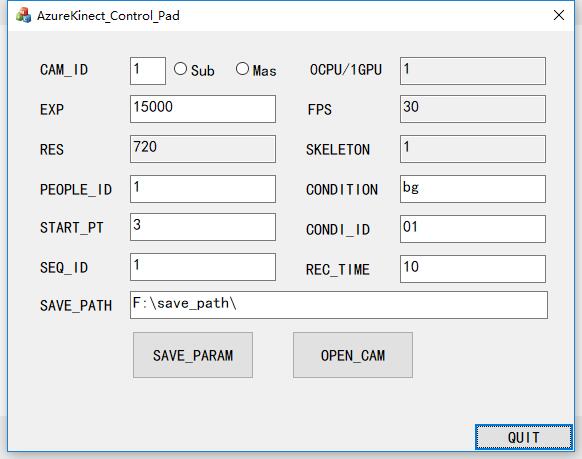}
\end{center}
\caption{The acquisition software we designed.}
\label{fig3}
\end{figure}

\textbf { Acquisition software } As shown in Fig.\ref{fig3}, we design an acquisition software based on the Developer Kit provided by Microsoft. During the acquisition process, the master device sends out the acquisition signal, and then the three cameras synchronously capture RGB+D video in MJPEG format and our synchronization error reached 0 ms.

\begin{figure}[htb]
\begin{center}

\includegraphics[width=0.7\linewidth]{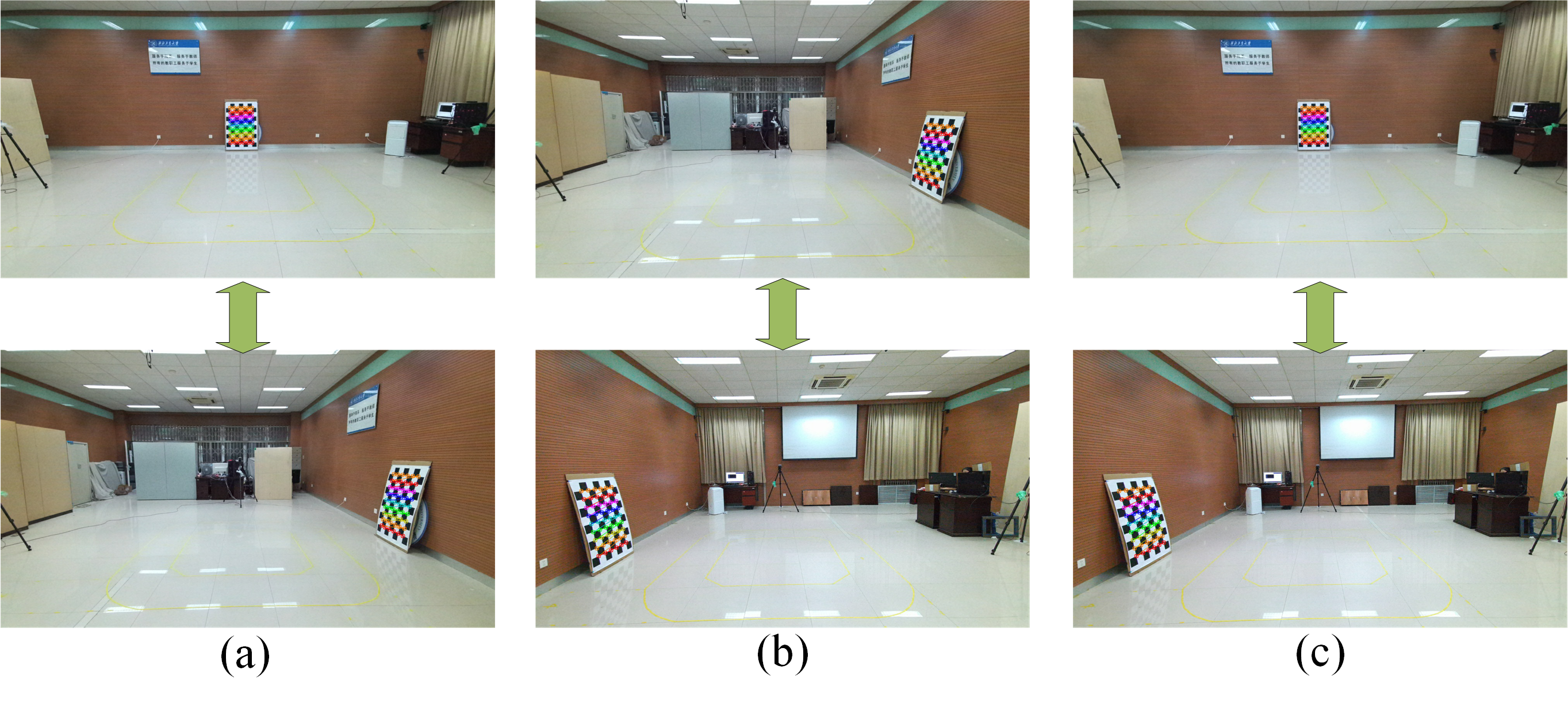}
\end{center}
\caption{Multiple binocular stereo calibrations.(a)binocular stereo calibration between subordinate device1 and  master device.(b) binocular stereo calibration between subordinate device1 and  subordinate device2.(c) binocular stereo calibration between subordinate device2 and master device.}
\label{fig4}
\end{figure}

\textbf {Multi-view 3D joints data fusion processing }Fusion of joints estimated from depth data obtained by multiple devices can effectively optimize some inaccurate joints collected by monocular camera due to occlusion. However, the coordinate systems of depth camera (D-CSYS) of each device are different and cannot be calibrated directly. Therefore, the joints data collected by different devices cannot be directly fused. To solve this problem, we align the joints data collected by the other two devices with the joints data collected by the target device under color camera coordinate system (C-CSYS) of the target device through multiple binocular stereo calibrations(as shown in Fig.\ref{fig4}), then fuse these aligned joints from multi-devices.

\begin{figure}[htb]
\begin{center}
\includegraphics[width=0.7\linewidth]{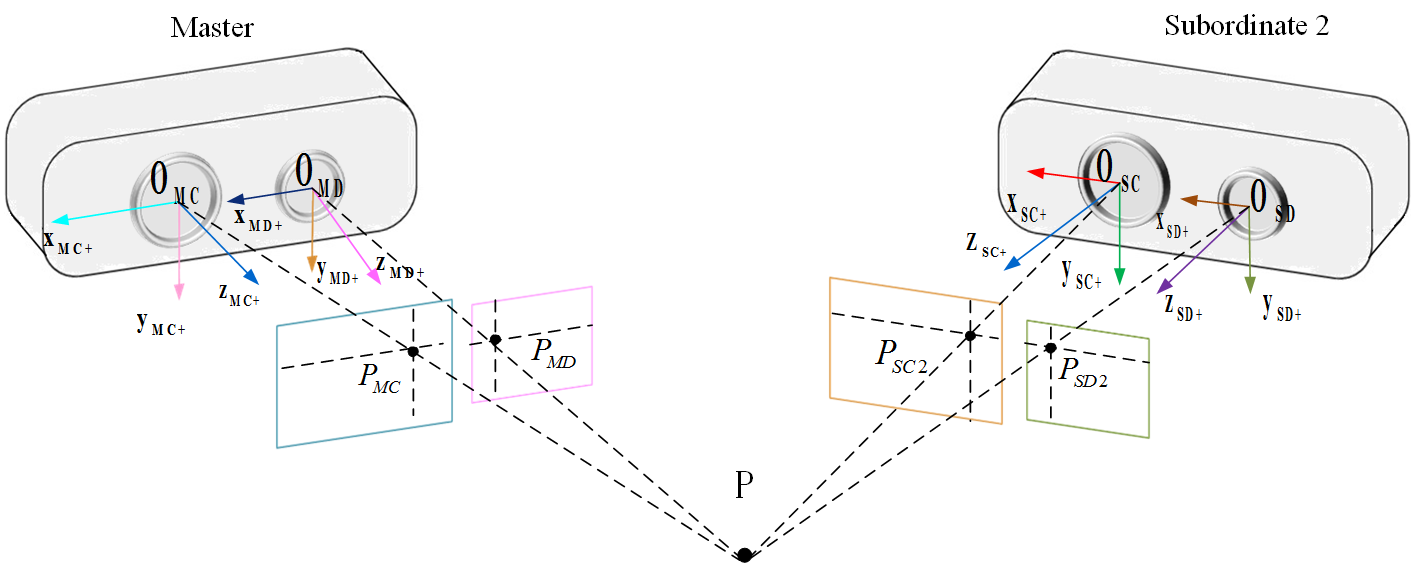}
\end{center}
\caption{The schematic diagram of data acquisition by the master device and subordinate device2.}
\label{fig5}
\end{figure}

For example, as shown in Fig. \ref{fig5}, regarding the master device as the target device, and we align the joints data captured by subordinate device2 with it. Suppose $P$ is a joint point of the subject, and ${{P}_{MC}}$ and ${{P}_{MD}}$ are the corresponding data collected by the color camera and the depth camera of the master device. ${{O}_{MC}}$ and ${{O}_{MD}}$ are the coordinate origin of the C-CSYS and D-CSYS of the master device, respectively. For each device, ${{O}_{MD}}$ of the D-CSYS is regarded as the coordinate origin of the world coordinate system. Then the rotation matrix ${{R}_{M}}$ and translation vector ${{T}_{M}}$ of the corresponding relation between the C-CSYS and the D-CSYS of the master device can be obtained through the binocular stereo calibration, besides ${{R}_{S2}}$ and ${{T}_{S2}}$ of subordinate device2 can be obtained in the same way. Thus ${{P}_{MC}}$ can be aligned with ${{P}_{MD}}$ by equation \eqref{eqMR}. Similarly, ${{P}_{SD2}}$ can be aligned with ${{P}_{SC2}}$ by equation \eqref{eqSR2}. Then, by binocular stereo calibration between C-CSYS of master device and C-CSYS of subordinate device2, ${P}_{SD2}$ can be aligned with ${{P}_{MC}}$ by equation \eqref{eqPM2}. Thus, by equation \eqref{eqMR} and \eqref{eqPM2}, ${P}_{SD2}$ can be aligned with ${{P}_{MD}}$ in C-CSYS of the target device.

\begin{equation}
{{P}_{MC}}={{R}_{M}}{{P}_{MD}}+{{T}_{M}}
\label{eqMR}
\end{equation}
\begin{equation}
{{P}_{SC2}}={{R}_{S2}}{{P}_{SD2}}+{{T}_{S2}}
\label{eqSR2}
\end{equation}
\begin{equation}
{{P}_{MC}}={{R}_{S2M}}{({{R}_{S2}}{{P}_{SD2}}+{{T}_{S2}})}+{{T}_{S2M}}+{{R}_{M}}{{T}_{M}}
\label{eqPM2}
\end{equation}
where ${{R}_{S2M}}$ and ${{T}_{S2M}}$ are the rotation matrix and translation vector between two C-CSYSs. And ${{R}_{M}}{{T}_{M}}$ is the offset between ${{O}_{MC}}$ and ${{O}_{MD}}$.

\begin{figure*}[htb]
\centering

\includegraphics[width=1\linewidth]{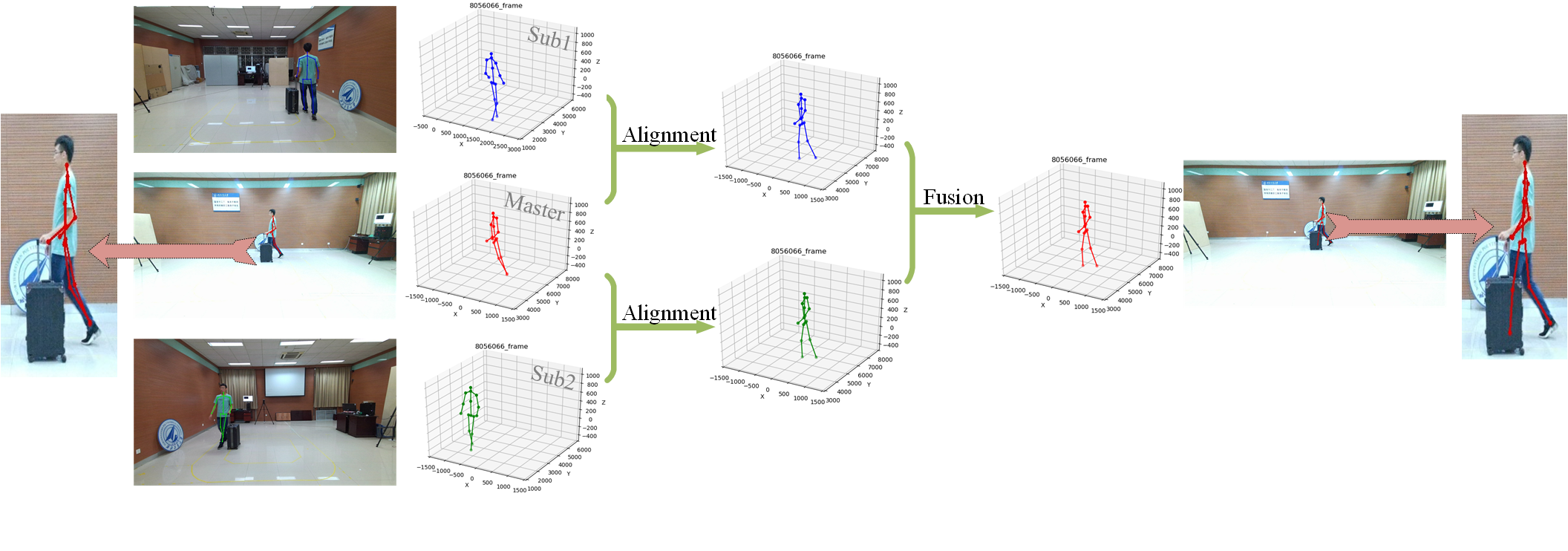}
\centering
\caption{An example of getting the optimized joints data. The master device is the target device. The C-CSYS of subordinate device1 and C-CSYS of subordinate device2 are respectively aligned with the C-CSYS of the master device through multiple binocular camera calibrations, the relatively accurate joints of two subordinate devices are fused with the inaccurate joints of the master device, which makes the joints of the master device more accurate, in this way, we get the optimized joints data of the master device.} 

\label{fig6}
\end{figure*}

For each device, if the data collected by the device is inaccurate due to occlusion, the optimized joints data can be obtained by fusing the aligned data of the other two devices. This is because there is at least one camera that can capture the front or back of the subject in the process of collecting data using multi-Kinect SDAS, which is less affected by self-occlusion. For example, as shown in Fig. \ref{fig6}, The joints collected by the master device are not accurate because of occlusion. After fusion with the accurate joints of the other two subordinate devices, the optimized joints of the master device can more accurately represent the human body.

\section{The OG RGB+D database}

The central motivation behind the OG RGB + D database is to support the study of gait recognition under occlusion, which is both available for model-based gait recognition methods and appearance-based ones. To meet this goal, we use multi-Kinect SDAS to collect multimodal single-gait and multi-gait data with various occlusion of 96 subjects (38\% female, 152 cm to 188 cm height, 40 kg to 104 kg weight, 19 years old to 54 years old and 85\% under 30 years old), including RGB+D data, optimized 3D joints data, and accurate silhouettes. In total, there are $96\times [7\text{ walking conditions (2LCL)}\times (12\times 4\text{ straight views+}9\times 4\text{ turning views})\text{+} 6\times 4\text{ multi-gait views}]\text{=}58752$ video clips ranged from 12 frames to 80 frames in OG RGB+D database.

\begin{figure}[htb]
\begin{center}
\includegraphics[width=0.89\linewidth]{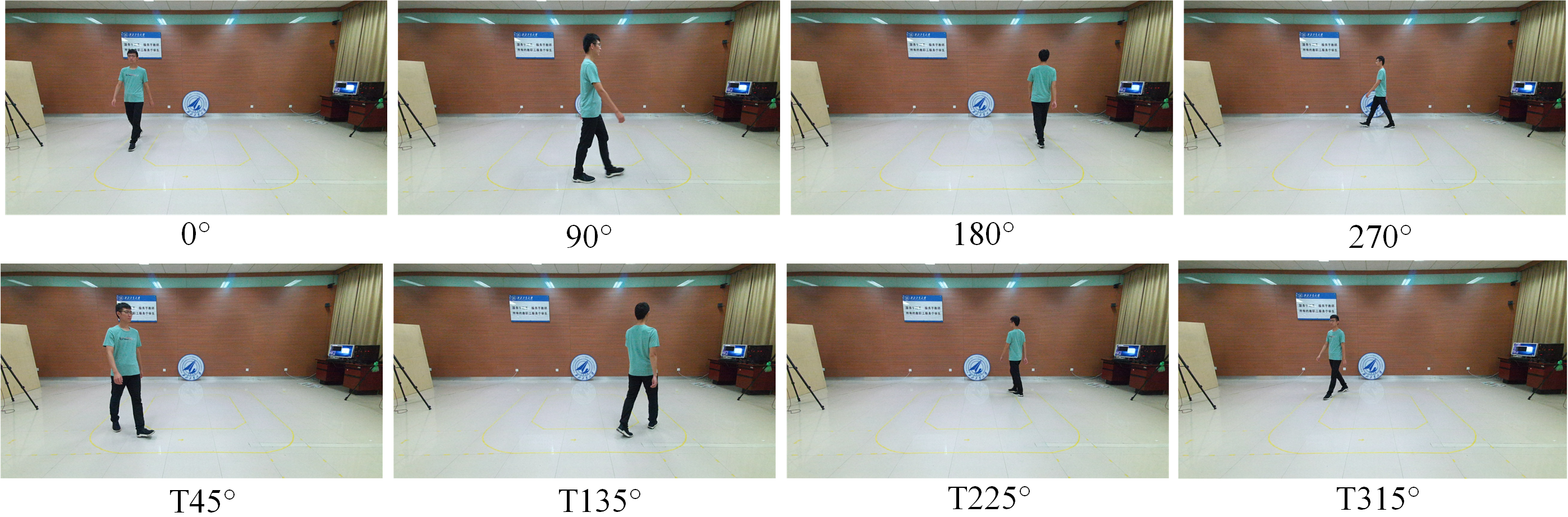}.
\end{center}
\caption{ Eight views of single-gait. Different views could cause different forms of self-occlusion.}
\label{fig7}
\end{figure}

\textbf{Self-occlusion} Different views could cause different forms of self-occlusion, which makes cross-view gait recognition more challenging. To acquire more views of gait in an undisturbed manner when the subject walks through the environment, we adopt a square route when collecting single-gait as shown in Fig. \ref{fig2}, which can collect the most challenging four straight views(0\degree, 90\degree, 180\degree, 270\degree) and four kinds of turning data (T45\degree, T135\degree, T225\degree, T315\degree) as shown in Fig. \ref{fig7}.

\begin{figure}[htb]
\begin{center}
\includegraphics[width=0.89\linewidth]{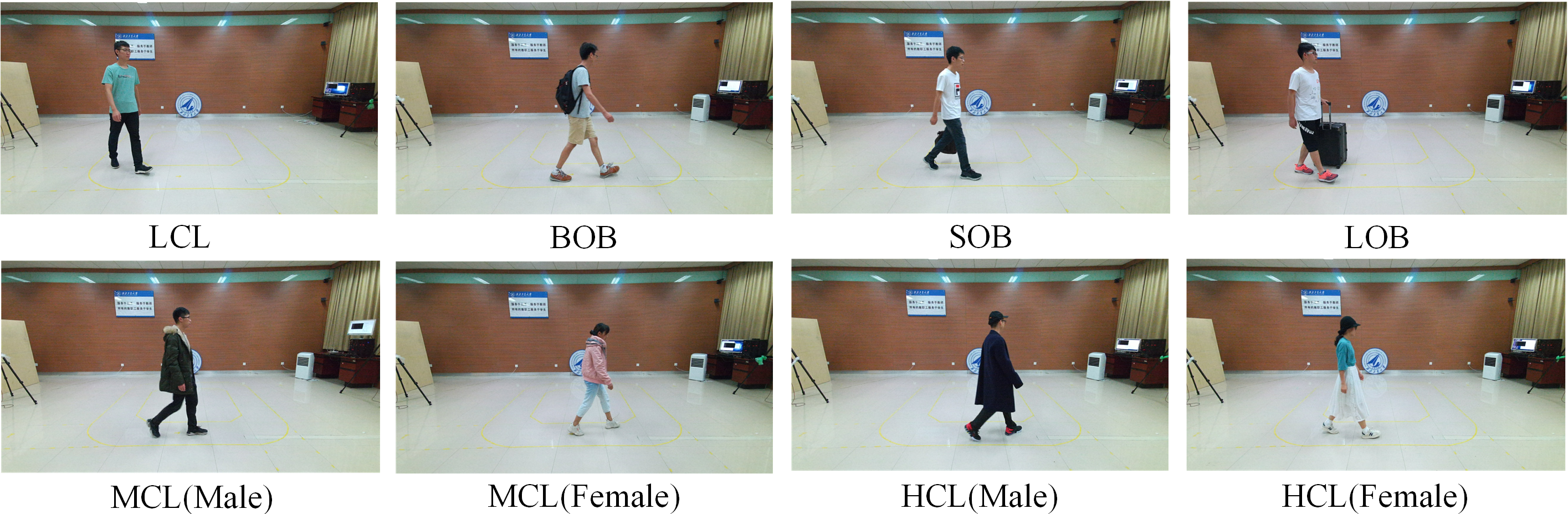}
\end{center}
\caption{Walking under active occlusion conditions. BOB: back object; SOB: side small object; LOB: side large object; LCL: low-occlusion clothing; MCL: medium-occlusion clothing and HCL: high-occlusion clothing.}
\label{fig8}
\end{figure}

\textbf{Active occlusion} In daily life, what people actively choose to carry or their some styles of clothes not only changes people's appearance, but also likely change their movement pattern, which increases the difficulty of gait recognition. As shown in Fig. \ref{fig8}, we collect six walking conditions respectively along with three common carried objects and three common styles of clothes according to the different degree of occlusions: back object (BOB), side small object (SOB), side large object (LOB), low-occlusion clothing (LCL), medium-occlusion clothing (MCL) and high-occlusion clothing (HCL). 

\begin{figure}[htb]
\begin{center}
\includegraphics[width=0.89\linewidth]{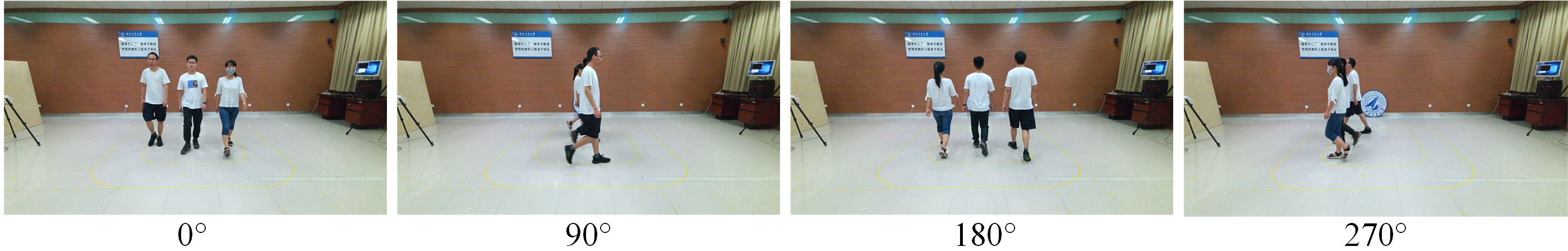}
\end{center}
\caption{ Four views of multi-gait (MG). At 90\degree and 270\degree, the passive occlusion is the most serious.}
\label{fig9}
\end{figure}

\textbf{Passive occlusion} When a group of people walk together, their gait may be occluded by each other, especially when their walking direction is vertical to the camera. For fully capturing this most extreme passive occlusion, three subjects are asked to walk straight when collecting multi-gait(as shown in Fig. \ref{fig2}(c)), in this way, we obtain four views of multi-gait (0\degree, 90\degree, 180\degree, 270\degree) as shown in Fig. \ref{fig9}.

\textbf{Silhouette extraction} The recognition accuracy of appearance-based algorithms depends heavily on silhouettes. To provide high-quality silhouettes for appearance-based methods and handle with complex background, illumination changes and shadows in experiment environment, we adopt the latest background matting technology \cite{ sengupta2020background} to extract silhouettes of single-gait, which can capture high quality foreground in natural settings, and we utilize pose-based human instance segmentation method Pose2Seg \cite{2019Pose2Seg} for multi-gait, which can moreover better handle occlusion by using human pose.


\section{ SkeletonGait}

\begin{figure*}
\begin{center}

\includegraphics[width=1\linewidth]{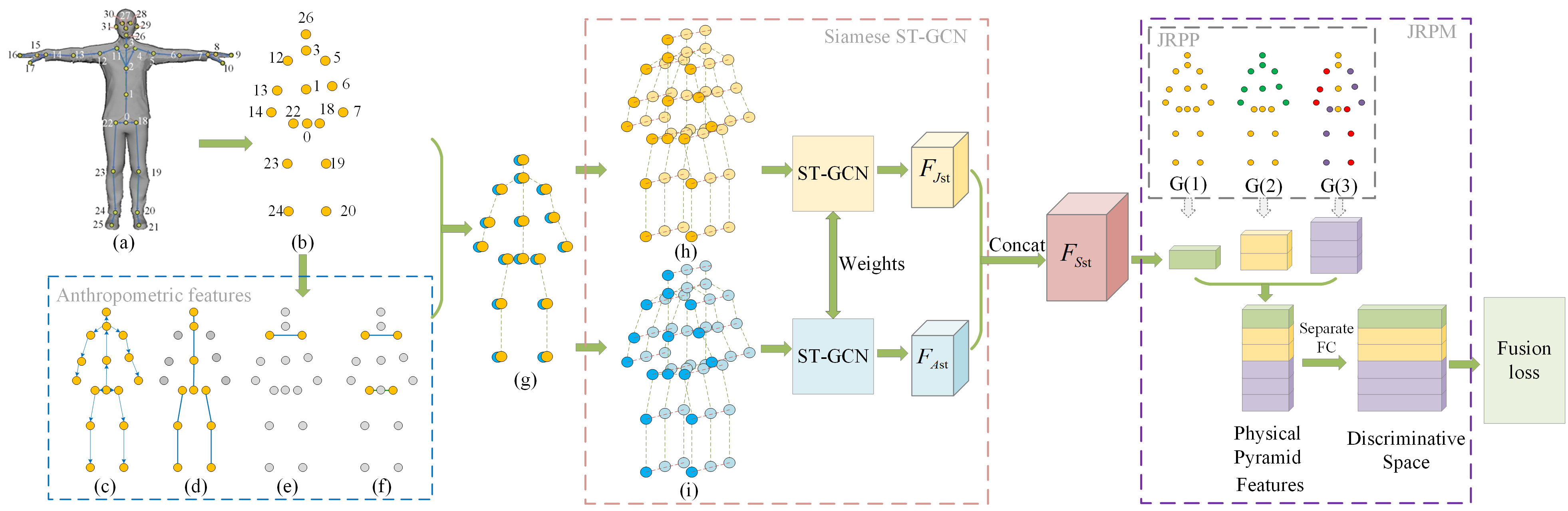}

\end{center}
\caption{ The overall pipeline of SkeletonGait. Selected human body 3D joints (b) from 32 joints (a) captured by Kinect DK and anthropometric features ((c) bones, (d) height, (e) shoulder breadth, (f) SHR) are used to build human dual skeleton model (g). By siamese ST-GCN framwork, the spatio-temporal gait features ${F}_{S\text{st}}$ consists of ${F}_{J\text{st}}$ and ${F}_{A\text{st}}$ are extracted from gait graphs (h) and (i), which are mapping into discriminative space by JRPM. Finally, a fusion loss strategy is applied to optimize our method.}
\label{fig10}
\end{figure*}

In this section, we introduce a novel model-based gait recognition method called SkeletonGait, which learns more discriminative gait information from human dual skeleton model we constructed using the siamese ST-GCN framwork. Especially, as Fig. \ref{fig10} shown, firstly, we calculate anthropometric features (Fig. \ref{fig10} (c)-(f)) from human 3D joints (Fig. \ref{fig10} (b)), and bond them to  joints to form a pseudo skeleton, thus we build a human dual skeleton model composed of the pseudo skeleton and the real skeleton (Fig. \ref{fig10} (g)) for better representing the human body. Then two spatio-temporal gait graphs (Fig. \ref{fig10} (h ), (i)) are made by following the work of ST-GCN \cite{yan2018spatial} for modeling the structured information when people are walking, which are fed into a siamese ST-GCN framework to extract different levels of spatio-temporal gait features, respectively. Finally, we further extract physical pyramid features from these features for enhancing local features and map them into a more discriminative space for gait recognition by our prior work, Joints Relationship Pyramid Mapping (JRPM) method \cite{Li2020}. Besides, we use a fusion loss strategy to handle variations better. The following parts of this section present implementation details.

\subsection{ Human Dual Skeleton Model }

For effectively representing the human body, we design a human dual skeleton model including the real human skeleton composed of joints and the pseudo skeleton composed of anthropometric features (bones, height, shoulder breadth (SB) and shoulder-to-hip ratio (SHR)).

Kinect DK can capture 32 joints(Fig. \ref{fig10} (a)), and we ignore the joints that have little effect on gait or difficult to estimate in our experiments, such as the joints of face, hands and feet, and select 16 important joints (Fig. \ref{fig10} (b)). 

We define that bones flowing from the center of the body to the extremities. Each bone links the parent joint with a child joint and is represented as a vector. For example, given a bone with its parent joint ${{\mathbf{j}}_{\mathbf{1}}}=({{x}_{1}},{{y}_{1}},{{z}_{1}})$and its child joint ${{\mathbf{j}}_{\mathbf{2}}}=({{x}_{2}},{{y}_{2}},{{z}_{2}})$, the vector of the bone is calculated as ${{\mathbf{b}}_{\mathbf{j1,j2}}}=({{x}_{1}}-{{x}_{2}},{{y}_{1}}-{{y}_{2}},{{z}_{1}}-{{z}_{2}})$.

Human height, as one of the most important anthropometric features, could be used to recognize gait \cite{yang2016relative}. For a walking person,  measuring its true height is difficult due to the posture difference. To solve this problem, as shown in Fig. \ref{fig10} (d) and equation \eqref{eqH}, the person’s height is calculated as the sum of the neck length, upper and lower spine lengths, and the average lengths of the right and left thighs and shanks.

\begin{equation}
\begin{split}
&H=\left| {{\mathbf{b}}_{j3,j26}} \right|+\left| {{\mathbf{b}}_{j1,j3}} \right|+\left| {{\mathbf{b}}_{j0,j1}} \right|\\
&+\frac{\left| {{\mathbf{b}}_{j22,j23}} \right|+\left| {{\mathbf{b}}_{j23,j24}} \right|+\left| {{\mathbf{b}}_{j18,j19}} \right|+\left| {{\mathbf{b}}_{j19,j20}} \right|}{2}
\end{split}
\label{eqH}
\end{equation}

Since biological studies \cite{kozlowski1978recognizing} have demonstrated the effect of SHR on gender classification , we take $SB=\left| {{\mathbf{j}}_{12}}-{{\mathbf{j}}_{5}} \right|$ and $SHR=\frac{\left| {{\mathbf{j}}_{12}}-{{\mathbf{j}}_{5}} \right|}{\left| {{\mathbf{j}}_{22}}-{{\mathbf{j}}_{18}} \right|}$ into account.

Then we assign these anthropometric features to form a pseudo skeleton. Each bone is assigned with a unique child joint. Because the number of bones is one less than the number of joints and joint 0 is not assigned to any bones, we add a fake bone with its value as $(H,SB,SHR)$ to joint 0. In this way, the pseudo skeleton and the real skeleton are the same in structure, and the human dual skeleton model with more body information is constructed.

\subsection{Siamese ST-GCN}

Although siamese network framework is generally used to learn the similarity of two different input data, we tentatively design siamese ST-GCN to extract spatio-temporal gait features from human dual skeleton model, which use ST-GCN \cite{yan2018spatial} as backbone. Because the real human skeleton composed of joints and the pseudo skeleton composed of human anthropometric features both reflect the gait information of the same person, even if they contain different body information. And we hope that these  different information could influence each other, so we share their network weights by adopting siamese network framework. As shown in Fig. \ref{fig10} (h), (i)), spatio-temporal gait graphs of human dual skeleton model are made respectively by following the work of ST-GCN \cite{yan2018spatial}, then they are fed into siamese ST-GCN framework to extract joint-level spatio-temporal gait features (${F}_{J\text{st}}$) and anthropometric spatio-temporal gait features (${F}_{A\text{st}}$), respectively. 

\begin{equation}
{{F}_{J\text{st}}}=(\sum\limits_{k}^{{{K}_{v}}}{{{\mathbf{W}}_{k}}({{\mathbf{f}}_{J}}\mathbf{\Lambda }_{k}^{-\tfrac{1}{2}}{{\mathbf{A}}_{k}}\mathbf{\Lambda }_{k}^{-\tfrac{1}{2}})})*{{K}_{Jt}}
\label{eqJ}
\end{equation}
\begin{equation}
{{F}_{A\text{st}}}=(\sum\limits_{k}^{{{K}_{v}}}{{{\mathbf{W}}_{k}}({{\mathbf{f}}_{A}}\mathbf{\Lambda }_{k}^{-\tfrac{1}{2}}{{\mathbf{A}}_{k}}\mathbf{\Lambda }_{k}^{-\tfrac{1}{2}})})*{{K}_{At}}
\label{eqA}
\end{equation}
where ${{\mathbf{f}}_{J}}$ and ${{\mathbf{f}}_{A}}$ are spatio-temporal gait graphs, ${{K}_{v}}$ is the kernel size of the spatial dimension, which is set to 3. ${{\mathbf{A}}_{k}}$ is the $N\times N$ adjacency matrix, whose element ${{\mathbf{A}_{k}^{ij}}}$ shows whether the vertex ${{v}_{j}}$ belongs to the subset ${{S}_{ik}}$ of vertex ${{v}_{i}}$.The normalized diagonal matrix $\mathbf{\Lambda} _{k}^{ii}=\sum\nolimits_{j}{(\mathbf{A}_{k}^{ij})}+\alpha $, where $\alpha \text{=}0.001$ for avoiding empty rows in ${{\mathbf{A}}_{k}}$.  ${{\mathbf{W}}_{k}}$ is the weight matrix.  ${{K}_{Jt}}$ and ${{K}_{At}}$ are the convolution kernels in temporal dimension.

By integrating ${F}_{J\text{st}}$ and ${F}_{A\text{st}}$, the final spatio-temporal gait can be obtained as
\begin{equation}
{{F}_{S\text{st}}}\text{=concat (}{{F}_{J\text{st}}},{{F}_{A\text{st}}}\text{)}
\label{eqST}
\end{equation}

\subsection{ Joints Relationship Pyramid Mapping}

In our prior work \cite{Li2020}, JRPM was designed to map spatio-temporal gait features into a more discriminative space through enhancing the discriminative gait feature of partial human body, which consists of Joints Relationship Pyramid Pooling (JRPP) and separate fully connect layers (FC). 

JRPP has $S$ scales. On scale $s\in 1,2,...,S$, the spatio-temporal gait features ${F}_{S\text{st}}$ is split into ${{G}_{s}}$ groups local features, and there are $\sum\limits_{s=1}^{S}{\sum\limits_{g=1}^{{{G}_{S}}}{{{F}_{s,g}}}}$ in total, where ${{F}_{s,g}}$ is a local feature, $s$ denotes the index of scale and $g$ denotes the index of local feature in each scale. For instance, ${{F}_{3,1}}$ means the first local feature in third scale. According to our prior research results \cite{Li2020}, the accuracy of gait recognition is highest on 3 pyramid scales. Hence, we finally adopt 3 pyramid scales within JRPP as shown in Fig. \ref{fig10}. Specifically, the whole body is regard as a group in G(1); According to horizontal plane in anatomy, The upper and lower body are treated as two groups in G(2); In G(3), left arm and right leg are regard as a group according to human general walking habit, and same for right arm and left leg, because people always naturally swing their arms when they walking, which is exactly opposite to the direction of their legs.
Then, each local feature ${{F}_{s,g}}$ is pooled by convolution kernel ${{k}_{s,g}}$ to generate physical pyramid feature ${{F}_{PP}}$:

\begin{equation}
{{F}_{PP}}=\sum\limits_{s=1}^{S}{\sum\limits_{g=1}^{{{G}_{S}}}{{{F}_{s,g}}}}*{{k}_{s,g}} 
\label{eqPP}
\end{equation}
where the size of ${{F}_{s,g}}$ is $N\times C\times {{J}_{s,g}}\times T$, ${{J}_{s,g}}$ is the number of joints in local body and the size of ${{k}_{s,g}}$ is ${{J}_{i,j}}\times T$.

Independent fully connect layers are used to map  ${{F}_{PP}}$ into the discriminative space to obtain the final gait feature representation. In such a way, the discernment of human gait could be obtained from coarse to fine.

\subsection{ Fusion loss }

Thus far, the gait features we extracted contain both spatio-temporal information and physiological characteristics. However, different views of the same identity usually have huge visual differences because of the influence of views. To solve this problem, we design a fusion loss function as 
\begin{equation}
L=\lambda {{L}_{tri\_BH}}+(1-\lambda ){{L}_{arc}}
\label{eq8}
\end{equation}
where $L_tri\_BH$ is improved triplet loss followed by \cite{hermans2017defense} that is used to increase the distance between different identity at same views, and reduce the distance between same identity at different views. And ${{L}_{arc}}$ is the arcface loss \cite{deng2019arcface} that is used to further increase inter-class distance. $\lambda $ is set to 0.9, which is the best value in our experiments.

\section{ Experiments and discussion }

Firstly, we define the benchmark evaluation of the OG RGB+D database. Secondly, we conduct ablation experiments to evaluate the effectiveness of each part of SkeletonGait and the accuracy of the optimized joints data. Next, we evaluate state-of-the-art gait recognition methods and SkeletonGait on our database to verify the performance of these methods in gait recognition under occlusion. Then we further evaluate the performance of SkeletonGait on public gait database CASIA-B and the effectiveness of the OJ data in improving the 3D pose estimation model. Finally, we compare the performance of gait recognition methods on the OG RGB+D database and the CASIA-B database to prove the significance of constructing OG RGB+D database.

\subsection{ Benchmark evaluations }
To have standard evaluations for the methods to be tested on this benchmark, we divide 96 subjects into two groups, each containing equal numbers of females and males. The training set consists of all the data of 48 subjects. In the test set, the gallery set is the remaining 48 subjects' gait data under LCL condition, and the probe set contains six walking conditions (BOB, SOB, LOB, MCL, HCL, MG). For each experimental setting, we report the recognition accuracy in percentage.

\subsection{ Ablation experiments}
\label{ablation}
 
Table \ref{table1} shows the results of ablation experiments.

\begin{table*}[htb]
\caption{Ablation Experiments on OG RGB+D database.}
\centering

\scalebox{0.85}{
\begin{tabular}{c|c|cccccccccc|c}
\hline\hline
\multirow{3}{*}{Input} & \multirow{3}{*}{Strategy} & \multicolumn{11}{c}{Accuracy (\%)}                                                                                                                                                                \\ \cline{3-13} 
                       &                           & \multicolumn{4}{c|}{OB}                                                           & \multicolumn{3}{c|}{CL}                               & \multicolumn{4}{c}{MG}                                                           \\ \cline{3-13} 
                       &                           & \multicolumn{1}{c|}{BOB}  & \multicolumn{1}{c|}{SOB}  & \multicolumn{1}{c|}{LOB}& \multicolumn{1}{c|}{\textbf{Mean}}  & \multicolumn{1}{c|}{MCL}  & \multicolumn{1}{c|}{HCL}& \multicolumn{1}{c|}{\textbf{Mean}}   & \multicolumn{1}{c|}{S}    & \multicolumn{1}{c|}{D}    & \multicolumn{1}{c|}{T}  & \multicolumn{1}{c}{\textbf{Mean}}   \\ \hline
\multirow{2}{*}{D-S(OJ)}   & ST-GCN\cite{yan2018spatial}(baseline) & \multicolumn{1}{c|}{17.2} & \multicolumn{1}{c|}{15.9} & \multicolumn{1}{c|}{14.1}  & \multicolumn{1}{c|}{15.7} & \multicolumn{1}{c|}{10.1} & \multicolumn{1}{c|}{9.7} & \multicolumn{1}{c|}{9.9} & \multicolumn{1}{c|}{11.3} & \multicolumn{1}{c|}{2.5} & \multicolumn{1}{c|}{1.2} & \multicolumn{1}{c}{5.0}\\
&\textbf{ siamese ST-GCN+JRPM} &\multicolumn{1}{c|}{\textbf{48.2}} & \multicolumn{1}{c|}{\textbf{44.8}} & \multicolumn{1}{c|}{\textbf{41.0}}&\multicolumn{1}{c|}{\textbf{44.7}} & \multicolumn{1}{c|}{\textbf{33.1}} & \multicolumn{1}{c|}{\textbf{30.0}} &\multicolumn{1}{c|}{\textbf{31.6}}& \multicolumn{1}{c|}{\textbf{33.4}} & \multicolumn{1}{c|}{\textbf{12.5}} & \multicolumn{1}{c|}{\textbf{6.0}}& \multicolumn{1}{c}{\textbf{17.3}} \\ 
S-S(OJ)
& ST-GCN+JRPM &\multicolumn{1}{c|}{30.0} & \multicolumn{1}{c|}{30.3} & \multicolumn{1}{c|}{29.0}&\multicolumn{1}{c|}{29.8} & \multicolumn{1}{c|}{23.7} & \multicolumn{1}{c|}{25.8} &\multicolumn{1}{c|}{24.8}& \multicolumn{1}{c|}{22.0} & \multicolumn{1}{c|}{5.9} & \multicolumn{1}{c|}{2.5} & \multicolumn{1}{c}{10.1}\\ 

D-S(J)
 &siamese ST-GCN+JRPM & \multicolumn{1}{c|}{24.9} & \multicolumn{1}{c|}{23.8} & \multicolumn{1}{c|}{20.5}&\multicolumn{1}{c|}{23.1} & \multicolumn{1}{c|}{16.9} & \multicolumn{1}{c|}{17.8} &\multicolumn{1}{c|}{17.4}& \multicolumn{1}{c|}{18.2} & \multicolumn{1}{c|}{3.3} & \multicolumn{1}{c|}{1.7}& \multicolumn{1}{c}{7.7}\\ 
D-S(VIBE\cite{2020VIBE})
& siamese ST-GCN+JRPM &\multicolumn{1}{c|}{10.2} & \multicolumn{1}{c|}{8.9} & \multicolumn{1}{c|}{6.3}&\multicolumn{1}{c|}{8.5} & \multicolumn{1}{c|}{5.0} & \multicolumn{1}{c|}{4.6}&\multicolumn{1}{c|}{4.8} & \multicolumn{1}{c|}{5.2} & \multicolumn{1}{c|}{1.0} & \multicolumn{1}{c|}{0.4}& \multicolumn{1}{c}{2.2} \\ 

D-S(work\cite{moon2019camera})
 &siamese ST-GCN+JRPM & \multicolumn{1}{c|}{22.3} & \multicolumn{1}{c|}{21.2} & \multicolumn{1}{c|}{19.9}&\multicolumn{1}{c|}{21.1} & \multicolumn{1}{c|}{13.1} & \multicolumn{1}{c|}{15.8}&\multicolumn{1}{c|}{14.5} & \multicolumn{1}{c|}{16.7} & \multicolumn{1}{c|}{2.3} & \multicolumn{1}{c|}{1.0}& \multicolumn{1}{c}{6.7}\\  \hline\hline
\multicolumn{13}{p{160mm}}{D-S: dual skeleton model;  S-S: single skeleton model;  OJ: optimized joints;  J: raw joints collected by Kinect;  VIBE\cite{2020VIBE}: joints estimated by VIBE\cite{2020VIBE};  work\cite{moon2019camera}: joints estimated by work\cite{moon2019camera};  S: single-gait;  D: double-gait;  T: triple-gait.}

\end{tabular}

}
\centering

\label{table1}
\end{table*}

\textbf{Baseline} The spatio-temporal features extracted by the baseline ST-GCN \cite{yan2018spatial} designed for action classification can reflect the motion information of walking, but its performance is not satisfactory for gait recognition task as shown in Table \ref{table1}. With the help of JRPM, the accuracy of gait recognition had been significantly improved by enhancing local features according to physiological characteristics.

\textbf{Human dual skeleton model} Table \ref{table1} shows that extracting gait features from human dual skeleton model made the accuracy of gait recognition effectively improve under all occlusion conditions. We can judge from experimental results that human dual skeleton model can more effectively represent the human body than single human skeleton model.

\textbf{OJ} The accuracy of gait recognition using the OJ data as input got marked improvement than that using raw joints obtained by Kinect DK as well as that using estimated joints by advanced 3D pose estimation method VIBE\cite{2020VIBE}and work \cite{moon2019camera}. Among them, the accuracy of gait recognition using the joints estimated by VIBE\cite{2020VIBE} was the lowest, because these joints are root relative and lose important location information, which lead to the loss of kinematic information of gait, such as step length. However, the joints estimated by work \cite {moon2019camera} are camera-centric, although the performance of gait recognition using work \cite{moon2019camera} was not as good as that using raw joints, its performance was relative satisfactory. Therefore, we use this model of work \cite{moon2019camera} to extract the joints of other gait recognition databases after training it on the OJ data of OG RGB+D database. After all, compared with raw joints, the OJ data could improve the accuracy by 21.6\% (from 23.1\% to 44.7\%) in OB, 14.2\% (from 17.4\% to 31.6\%) in CL, and 9.4\% (from 7.7\% to 17.3\%) in MG, respectively. This is because the optimized joints are more accurate with the help of our multi-Kinect SDAS and fusion strategy, so that the human dual skeleton model also can better represent the human body and the accuracy of gait recognition is higher.

\subsection{ Evaluation results on OG RGB+D database}
Table \ref{table2} shows that evaluation results of state-of-art model-based methods PoseGait\cite{liao2020model}, JointsGait\cite{Li2020}, appearance-based methods  Gaitset\cite{chao2019gaitset}, GaitPart\cite{fan2020gaitpart}, MT3D\cite{lin2020gait} and SkeletonGait on our database. And we train these state-of-art models according to their literatures, respectively.

\begin{table}[]
\caption{Evaluation of state-of-arts and SkeletonGait on OG RGB+D database, excluding identical-view cases. In each view, the highest accuracy of each type of methods is reported in \textbf{bold} and highest accuracy of all methods is  reported in \textbf{\textcolor[RGB]{253,104,100}{orange}}.}
\centering
\scalebox{0.9}{
\begin{tabular}{p{2mm}p{4mm}|c|cccccccc|c}
\hline\hline
& \multicolumn{2}{c|}{} & \multicolumn{9}{c}{Accuracy(\%)} \\ \cline{4-12} 
& \multicolumn{2}{c|}{} & \multicolumn{8}{c|}{0\degree,T45\degree,90\degree,T135\degree,180\degree,T225\degree,270\degree,T315\degree} & \multirow{2}{*}{\begin{tabular}[c]{@{}c@{}}Average\end{tabular}} \\ \cline{1-11}
& \multicolumn{2}{c|}{Probe} & 0\degree & T45\degree & 90\degree & T135\degree & 180\degree & T225\degree & 270\degree & T315\degree & \\ \hline

\multicolumn{1}{c|}{\multirow{18}{*}{OB}} & \multirow{6}{*}{BOB} & Gaitset\cite{chao2019gaitset}      & 61.7  & 71.3 & 54.2  &  70.5  &  58.6  &  67.9  &  50.6  &  71.0 & 63.2                                                   \\
\multicolumn{1}{c|}{}                     &                     & GaitPart\cite{fan2020gaitpart}     & 73.1  & 79.9   & 69.4 &	75.6 &	69.0 &	77.6 &	61.1 &	74.4 & 72.5                                                           \\
\multicolumn{1}{c|}{}                     &                      & MT3D\cite{lin2020gait}             & \textbf{{\color[HTML]{FD6864}81.5}}	& \textbf{{\color[HTML]{FD6864}89.9}}	& \textbf{{\color[HTML]{FD6864}79.8}}	& \textbf{{\color[HTML]{FD6864}83.8}}	& \textbf{{\color[HTML]{FD6864}76.4}}	& \textbf{{\color[HTML]{FD6864}83.9}}
	& \textbf{{\color[HTML]{FD6864} 64.0}}	& \textbf{{\color[HTML]{FD6864} 83.1}}	& \textbf{{\color[HTML]{FD6864} 80.3}}                                        \\ \cline{3-12} 

\multicolumn{1}{c|}{}                     &                     & \cellcolor[HTML]{D1DED0}PoseGait\cite{liao2020model}    & \cellcolor[HTML]{D1DED0}12.3	&\cellcolor[HTML]{D1DED0}11.3	&\cellcolor[HTML]{D1DED0}7.5	& \cellcolor[HTML]{D1DED0}14.3	& \cellcolor[HTML]{D1DED0}14.1	& \cellcolor[HTML]{D1DED0}13.4	& \cellcolor[HTML]{D1DED0}12.9	& \cellcolor[HTML]{D1DED0}15.8	& \cellcolor[HTML]{D1DED0}12.7                                                               \\
\multicolumn{1}{c|}{}                     &                     & \cellcolor[HTML]{D1DED0}JointsGait\cite{Li2020}          & \cellcolor[HTML]{D1DED0}16.9  & \cellcolor[HTML]{D1DED0}35.6  & \cellcolor[HTML]{D1DED0}16.7  & \cellcolor[HTML]{D1DED0}26.4  &\cellcolor[HTML]{D1DED0}20.6 &\cellcolor[HTML]{D1DED0}30.6 &\cellcolor[HTML]{D1DED0}22.3 &\cellcolor[HTML]{D1DED0}32.9 &\cellcolor[HTML]{D1DED0}25.3                                                     \\
\multicolumn{1}{c|}{}                     &                    & \cellcolor[HTML]{D1DED0}SkeletonGait                   &\cellcolor[HTML]{D1DED0}\textbf{47.2}  & \cellcolor[HTML]{D1DED0}\textbf{59.5}  &\cellcolor[HTML]{D1DED0} \textbf{43.0}  & \cellcolor[HTML]{D1DED0}\textbf{46.9}&	\cellcolor[HTML]{D1DED0}\textbf{42.1} &	\cellcolor[HTML]{D1DED0}\textbf{50.3} &\cellcolor[HTML]{D1DED0}\textbf{47.4} &\cellcolor[HTML]{D1DED0}\textbf{48.4}	 &\cellcolor[HTML]{D1DED0}\textbf{48.2}                                                  \\ \cline{2-12} 
\multicolumn{1}{c|}{}                     & \multirow{6}{*}{SOB} & Gaitset\cite{chao2019gaitset}      & 39.2  & 49.7  &37.6  & 54.9  &  31.5  &  38.4  &  20.6  &37.9 & 38.7                                                         \\
\multicolumn{1}{c|}{}                     &                      & GaitPart\cite{fan2020gaitpart}     & 59.7  & 66.2  &56.8   & 65.6 &	50.7 &	50.3 &	28.0 &	48.4 &53.2                                                          \\
\multicolumn{1}{c|}{}                     &                      & MT3D\cite{lin2020gait}             & \textbf{{\color[HTML]{FD6864}68.6}}	&\textbf{{\color[HTML]{FD6864}71.4}} 	&\textbf{{\color[HTML]{FD6864}58.9}}	&\textbf{{\color[HTML]{FD6864}72.4}}	&\textbf{{\color[HTML]{FD6864}58.5}} 	&\textbf{{\color[HTML]{FD6864}56.0}} 	&\textbf{33.2}	&\textbf{{\color[HTML]{FD6864}57.1}} 	& \textbf{{\color[HTML]{FD6864}59.5}} \\\cline{3-12}
\multicolumn{1}{c|}{}                     &                     & \cellcolor[HTML]{D1DED0}PoseGait\cite{liao2020model}     & \cellcolor[HTML]{D1DED0}10.9	& \cellcolor[HTML]{D1DED0}12.0	&\cellcolor[HTML]{D1DED0}9.1	&\cellcolor[HTML]{D1DED0}17.7	&\cellcolor[HTML]{D1DED0}15.3	& \cellcolor[HTML]{D1DED0}14.4	&\cellcolor[HTML]{D1DED0}12.4	&\cellcolor[HTML]{D1DED0}15.0	&\cellcolor[HTML]{D1DED0}13.3                                                                \\
\multicolumn{1}{c|}{}                     &                     & \cellcolor[HTML]{D1DED0}JointsGait\cite{Li2020}          &\cellcolor[HTML]{D1DED0}15.5  & \cellcolor[HTML]{D1DED0}30.4  &\cellcolor[HTML]{D1DED0}16.8  & \cellcolor[HTML]{D1DED0}28.6 & \cellcolor[HTML]{D1DED0}21.8 &\cellcolor[HTML]{D1DED0}27.6 &\cellcolor[HTML]{D1DED0}21.9 & \cellcolor[HTML]{D1DED0}27.7 &\cellcolor[HTML]{D1DED0}23.8                                                            \\
\multicolumn{1}{c|}{}                     &                   &\cellcolor[HTML]{D1DED0} SkeletonGait                &\cellcolor[HTML]{D1DED0}\textbf{39.2} &	\cellcolor[HTML]{D1DED0}\textbf{51.3} &\cellcolor[HTML]{D1DED0}	\textbf{41.1} &	\cellcolor[HTML]{D1DED0}\textbf{50.8} &\cellcolor[HTML]{D1DED0}\textbf{35.6} &\cellcolor[HTML]{D1DED0}\textbf{51.4} &\cellcolor[HTML]{D1DED0}	\textbf{{\color[HTML]{FD6864}37.5}} & \cellcolor[HTML]{D1DED0}\textbf{51.3} &\cellcolor[HTML]{D1DED0}\textbf{44.8}                                                   \\ \cline{2-12} 
\multicolumn{1}{c|}{}                     & \multirow{6}{*}{LOB} & Gaitset\cite{chao2019gaitset}      & 22.7  &  28.7  &  19.5  &  31.3	  &  25.7  &  20.6  &  8.4  &  15.5  & 21.5                                                \\
\multicolumn{1}{c|}{}                     &                      & GaitPart\cite{fan2020gaitpart}     & 37.1 &	44.2 &	29.9 &	\textbf{{\color[HTML]{FD6864}46.4}} &	46.0 &	31.8 &	9.9 &	20.9 & 33.2                                                           \\
\multicolumn{1}{c|}{}                     &                      & MT3D\cite{lin2020gait}            &\textbf{{\color[HTML]{FD6864}38.8}} 	& 38.3	& 32.2	& 42.2	&\textbf{{\color[HTML]{FD6864}47.3}} 	& \textbf{37.7} 	&\textbf {12.3}	& \textbf{26.3}	&\textbf{34.4}                           \\\cline{3-12}
\multicolumn{1}{c|}{}                     &                     & \cellcolor[HTML]{D1DED0}PoseGait\cite{liao2020model}     &\cellcolor[HTML]{D1DED0}11.4	& \cellcolor[HTML]{D1DED0}9.4	&\cellcolor[HTML]{D1DED0}6.9	& \cellcolor[HTML]{D1DED0}14.8	& \cellcolor[HTML]{D1DED0}12.7	& \cellcolor[HTML]{D1DED0}12.7	& \cellcolor[HTML]{D1DED0}10.7	&\cellcolor[HTML]{D1DED0}15.4	&\cellcolor[HTML]{D1DED0}11.8                                                                \\
\multicolumn{1}{c|}{}                     &                     &\cellcolor[HTML]{D1DED0} JointsGait\cite{Li2020}          & \cellcolor[HTML]{D1DED0}19.9  &\cellcolor[HTML]{D1DED0}28.4  &\cellcolor[HTML]{D1DED0}23.7  & \cellcolor[HTML]{D1DED0}23.9 & \cellcolor[HTML]{D1DED0}21.1 &\cellcolor[HTML]{D1DED0}25.8 & \cellcolor[HTML]{D1DED0}20.5 &\cellcolor[HTML]{D1DED0}24.9 &\cellcolor[HTML]{D1DED0}23.5                                                             \\
\multicolumn{1}{c|}{}                     &                    & \cellcolor[HTML]{D1DED0}SkeletonGait             & \cellcolor[HTML]{D1DED0}\textbf{35.0} & \cellcolor[HTML]{D1DED0}\textbf{{\color[HTML]{FD6864}47.9}} &\cellcolor[HTML]{D1DED0}\textbf{{\color[HTML]{FD6864}38.2}}&\cellcolor[HTML]{D1DED0}\textbf{44.0} & \cellcolor[HTML]{D1DED0}\textbf{34.9} & \cellcolor[HTML]{D1DED0}\textbf{{\color[HTML]{FD6864}45.2}}&\cellcolor[HTML]{D1DED0}\textbf{{\color[HTML]{FD6864}37.5}} & \cellcolor[HTML]{D1DED0}\textbf{{\color[HTML]{FD6864}45.1}} &\cellcolor[HTML]{D1DED0}\textbf{{\color[HTML]{FD6864}41.0}}\\\cline{1-12}
\multicolumn{1}{c|}{\multirow{12}{*}{CL}} & \multirow{6}{*}{MCL} & Gaitset\cite{chao2019gaitset}      & 34.1  &  36.1  &  30.4  &  32.7  &  27.9  &  39.2  &  29.7  &  42.4 & 34.0                                                      \\ 
\multicolumn{1}{c|}{}                     &                      & GaitPart\cite{fan2020gaitpart}     & 39.9 &	45.2 &	39.5 &	38.4 &	33.8 &	39.9 &	36.2 &	46.4 &39.9                                                               \\
\multicolumn{1}{c|}{}                     &                      & MT3D\cite{lin2020gait}             & \textbf{{\color[HTML]{FD6864}53.9}}	&\textbf{{\color[HTML]{FD6864}57.1}} 	&\textbf{{\color[HTML]{FD6864}49.9}} 	&\textbf{{\color[HTML]{FD6864}50.7}}	&\textbf{{\color[HTML]{FD6864}43.5}} &\textbf{{\color[HTML]{FD6864}53.5}}& \textbf{{\color[HTML]{FD6864}48.7}}& \textbf{{\color[HTML]{FD6864}59.2}}&\textbf{{\color[HTML]{FD6864}52.1}}                                            \\\cline{3-12}
\multicolumn{1}{c|}{}                     &                     & \cellcolor[HTML]{D1DED0}PoseGait\cite{liao2020model}     & \cellcolor[HTML]{D1DED0}11.2	& \cellcolor[HTML]{D1DED0}8.7	& \cellcolor[HTML]{D1DED0}7.0	&\cellcolor[HTML]{D1DED0}12.3	& \cellcolor[HTML]{D1DED0}10.4	& \cellcolor[HTML]{D1DED0}9.3	& \cellcolor[HTML]{D1DED0}9.8	& \cellcolor[HTML]{D1DED0}11.7	& \cellcolor[HTML]{D1DED0}10.1                                                                \\
\multicolumn{1}{c|}{}                     &                     & \cellcolor[HTML]{D1DED0}JointsGait\cite{Li2020}          & \cellcolor[HTML]{D1DED0}17.9  &\cellcolor[HTML]{D1DED0}30.6  & \cellcolor[HTML]{D1DED0}21.2  &\cellcolor[HTML]{D1DED0}23.9 & \cellcolor[HTML]{D1DED0}17.4 & \cellcolor[HTML]{D1DED0}22.5 & \cellcolor[HTML]{D1DED0}17.9 &\cellcolor[HTML]{D1DED0}24.9 &\cellcolor[HTML]{D1DED0}22.0                                                                \\
\multicolumn{1}{c|}{}                     &                    & \cellcolor[HTML]{D1DED0}SkeletonGait              & \cellcolor[HTML]{D1DED0}34.0  & \cellcolor[HTML]{D1DED0}\textbf{39.5}  & \cellcolor[HTML]{D1DED0}\textbf{33.1}  & \cellcolor[HTML]{D1DED0}\textbf{31.2}& \cellcolor[HTML]{D1DED0}\textbf{25.1}& \cellcolor[HTML]{D1DED0}\textbf{35.8} & \cellcolor[HTML]{D1DED0}\textbf{29.8} & \cellcolor[HTML]{D1DED0}\textbf{36.5} & \cellcolor[HTML]{D1DED0}\textbf{33.1}                                                       \\\cline{2-12}
\multicolumn{1}{c|}{}                     & \multirow{6}{*}{HCL} & Gaitset\cite{chao2019gaitset}      &18.2  &21.0  &  17.3  &  17.1  &  15.2  &  22.2  &  17.9  &  24.1 & 19.1                                                      \\ 
\multicolumn{1}{c|}{}                     &                      & GaitPart\cite{fan2020gaitpart}     &23.4 &26.1   &21.9 &	23.4 &	22.7 &	25.8 &	23.6 &	28.1  & 24.3                                                          \\
\multicolumn{1}{c|}{}                     &                      & MT3D\cite{lin2020gait}             & \textbf{25.7}	& \textbf{28.0}	& \textbf{25.9}	&\textbf{24.0}	& \textbf{21.4}	& \textbf{25.5}	&\textbf{{\color[HTML]{FD6864} 27.9}} 	& \textbf{31.6}	& \textbf{26.3}                                 \\\cline{3-12}
\multicolumn{1}{c|}{}                     &                     & \cellcolor[HTML]{D1DED0}PoseGait\cite{liao2020model}     &\cellcolor[HTML]{D1DED0}7.2	& \cellcolor[HTML]{D1DED0}8.2	&\cellcolor[HTML]{D1DED0}6.9	&\cellcolor[HTML]{D1DED0}11.7	&\cellcolor[HTML]{D1DED0}9.2	& \cellcolor[HTML]{D1DED0}9.8	&\cellcolor[HTML]{D1DED0}8.0	&\cellcolor[HTML]{D1DED0}13.0	&\cellcolor[HTML]{D1DED0}9.3                                                                 \\
\multicolumn{1}{c|}{}                     &                     & \cellcolor[HTML]{D1DED0}JointsGait\cite{Li2020}          & \cellcolor[HTML]{D1DED0}22.8  &\cellcolor[HTML]{D1DED0}28.8  &\cellcolor[HTML]{D1DED0}19.0  &\cellcolor[HTML]{D1DED0}21.3 &\cellcolor[HTML]{D1DED0}16.4 &\cellcolor[HTML]{D1DED0}20.8 &\cellcolor[HTML]{D1DED0}16.2 & \cellcolor[HTML]{D1DED0}22.5 & \cellcolor[HTML]{D1DED0}20.9                                                               \\
\multicolumn{1}{c|}{}                     &                    &\cellcolor[HTML]{D1DED0}SkeletonGait             &\cellcolor[HTML]{D1DED0}\textbf{{\color[HTML]{FD6864}28.0}}   &\cellcolor[HTML]{D1DED0}\textbf{{\color[HTML]{FD6864} 39.3}}   &\cellcolor[HTML]{D1DED0}\textbf{{\color[HTML]{FD6864}27.8}}   &\cellcolor[HTML]{D1DED0}\textbf{{\color[HTML]{FD6864}27.8}}  &\cellcolor[HTML]{D1DED0}\textbf{{\color[HTML]{FD6864}21.5}}  &\cellcolor[HTML]{D1DED0}\textbf{{\color[HTML]{FD6864} 35.2}}  &\cellcolor[HTML]{D1DED0}\textbf{24.1}  &\cellcolor[HTML]{D1DED0}\textbf{{\color[HTML]{FD6864} 35.7}} &\cellcolor[HTML]{D1DED0}\textbf{{\color[HTML]{FD6864} 30.0}} \\ \cline{1-12}
\multicolumn{1}{c|}{\multirow{18}{*}{MG}} & \multirow{6}{*}{S}   & Gaitset\cite{chao2019gaitset}      & 35.9  & $\setminus$  & \textbf{18.1}  & $\setminus$ & 31.0  & $\setminus$ & \textbf{18.7 }& $\setminus$ & \textbf{25.9} \\ 
\multicolumn{1}{c|}{}                     &                      & GaitPart\cite{fan2020gaitpart}     &\textbf{{\color[HTML]{FD6864}37.0}}   & $\setminus$  & 15.7  & $\setminus$  & \textbf{32.6} & $\setminus$ & 14.9 & $\setminus$& 25.1                         \\
\multicolumn{1}{c|}{}                     &                      & MT3D\cite{lin2020gait}             & 32.7  & $\setminus$  &11.7  & $\setminus$ & 29.9 & $\setminus$ & 14.3 & $\setminus$ & 22.2                         \\\cline{3-12}
\multicolumn{1}{c|}{}                     &                     &\cellcolor[HTML]{D1DED0}PoseGait\cite{liao2020model}     & \cellcolor[HTML]{D1DED0}9.8  & \cellcolor[HTML]{D1DED0}$\setminus$  &\cellcolor[HTML]{D1DED0}11.5  & $\cellcolor[HTML]{D1DED0}\setminus$ & \cellcolor[HTML]{D1DED0}11.1 &\cellcolor[HTML]{D1DED0}$\setminus$  &\cellcolor[HTML]{D1DED0}8.4 & \cellcolor[HTML]{D1DED0}$\setminus$ & \cellcolor[HTML]{D1DED0}10.2                                         \\
\multicolumn{1}{c|}{}                     &                     & \cellcolor[HTML]{D1DED0}JointsGait\cite{Li2020}          & \cellcolor[HTML]{D1DED0}22.1  &\cellcolor[HTML]{D1DED0} $\cellcolor[HTML]{D1DED0}\setminus$  & \cellcolor[HTML]{D1DED0}19.1  &\cellcolor[HTML]{D1DED0} $\setminus$ & \cellcolor[HTML]{D1DED0}21.0 & \cellcolor[HTML]{D1DED0}$\setminus$ & \cellcolor[HTML]{D1DED0}17.9 & \cellcolor[HTML]{D1DED0}$\setminus$ & \cellcolor[HTML]{D1DED0}20.0                                    \\
\multicolumn{1}{c|}{}                     &                    & \cellcolor[HTML]{D1DED0}SkeletonGait)              & \cellcolor[HTML]{D1DED0}\textbf{32.9}  &\cellcolor[HTML]{D1DED0} $\cellcolor[HTML]{D1DED0}\setminus$  &\cellcolor[HTML]{D1DED0}\textbf{{\color[HTML]{FD6864}35.8}}& $\cellcolor[HTML]{D1DED0}\setminus$ &\cellcolor[HTML]{D1DED0}\textbf{{\color[HTML]{FD6864}32.7}}  & \cellcolor[HTML]{D1DED0}$\setminus$ &\cellcolor[HTML]{D1DED0}\textbf{{\color[HTML]{FD6864}32.2}} &\cellcolor[HTML]{D1DED0} $\setminus$ &\cellcolor[HTML]{D1DED0}\textbf{{\color[HTML]{FD6864}33.4}}  \\\cline{2-12}
\multicolumn{1}{c|}{}                     & \multirow{6}{*}{D}   & Gaitset\cite{chao2019gaitset}      & \textbf{{\color[HTML]{FD6864}17.3}} & $\setminus$  & \textbf{3.6}  & $\setminus$ & \textbf{{\color[HTML]{FD6864}14.4}}& $\setminus$ & 2.0 & $\setminus$ & \textbf{9.3}                                     \\ 
\multicolumn{1}{c|}{}                     &                      & GaitPart\cite{fan2020gaitpart}     & 16.7  & $\setminus$  & 2.5  &$\setminus$  & 13.0 & $\setminus$ & 1.6 & $\setminus$ & 8.5                                      \\
\multicolumn{1}{c|}{}                     &                      & MT3D\cite{lin2020gait}             & 12.7  & $\setminus$ & 1.0  & $\setminus$ & 9.9 & $\setminus$ & \textbf{2.1} & $\setminus$ & 6.4                          \\\cline{3-12}
\multicolumn{1}{c|}{}                     &                     &\cellcolor[HTML]{D1DED0} PoseGait\cite{liao2020model}     &\cellcolor[HTML]{D1DED0}1.2  & \cellcolor[HTML]{D1DED0}$\setminus$  &\cellcolor[HTML]{D1DED0}0.8  & $\cellcolor[HTML]{D1DED0}\setminus$ &\cellcolor[HTML]{D1DED0}1.1 & \cellcolor[HTML]{D1DED0}$\setminus$ &\cellcolor[HTML]{D1DED0}0.5 &\cellcolor[HTML]{D1DED0} $\setminus$ &\cellcolor[HTML]{D1DED0}0.9                                        \\
\multicolumn{1}{c|}{}                     &                     & \cellcolor[HTML]{D1DED0}JointsGait\cite{Li2020}          &\cellcolor[HTML]{D1DED0}5.5  & \cellcolor[HTML]{D1DED0}$\setminus$  &\cellcolor[HTML]{D1DED0}5.1  & $\cellcolor[HTML]{D1DED0}\setminus$ &\cellcolor[HTML]{D1DED0}6.9 &\cellcolor[HTML]{D1DED0} $\setminus$ &\cellcolor[HTML]{D1DED0}5.2    &\cellcolor[HTML]{D1DED0} $\setminus$ &\cellcolor[HTML]{D1DED0}5.7                                        \\
\multicolumn{1}{c|}{}                     &                    & \cellcolor[HTML]{D1DED0}SkeletonGait              & \cellcolor[HTML]{D1DED0}\textbf{12.4}  & \cellcolor[HTML]{D1DED0}$\setminus$  & \cellcolor[HTML]{D1DED0}\textbf{{\color[HTML]{FD6864}13.4}} & \cellcolor[HTML]{D1DED0}$\setminus$ &\cellcolor[HTML]{D1DED0}\textbf{13.2} &\cellcolor[HTML]{D1DED0} $\setminus$ &\cellcolor[HTML]{D1DED0}\textbf{{\color[HTML]{FD6864}10.9}}  &\cellcolor[HTML]{D1DED0} $\cellcolor[HTML]{D1DED0}\cellcolor[HTML]{D1DED0}\setminus$ &\cellcolor[HTML]{D1DED0}\textbf{{\color[HTML]{FD6864}12.5}}  \\\cline{2-12}
\multicolumn{1}{c|}{}                     & \multirow{6}{*}{T}   & Gaitset\cite{chao2019gaitset}      & \textbf{{\color[HTML]{FD6864}13.0}}  & $\setminus$  &\textbf{0.7}   & $\setminus$ &\textbf{{\color[HTML]{FD6864}9.4}} & $\setminus$ & 0.3 &$\setminus$ & \textbf{5.9}                                       \\ 
\multicolumn{1}{c|}{}                     &                      & GaitPart\cite{fan2020gaitpart}     & 10.2  & $\setminus$  & \textbf{0.7}  & $\setminus$ & 7.7 & $\setminus$ &\textbf{0.5}  &$\setminus$ & 4.8                                       \\
\multicolumn{1}{c|}{}                     &                      & MT3D\cite{lin2020gait}          & 6.4  & $\setminus$  & 0.0 &$\setminus$ & 3.0 &$\setminus$ &\textbf{0.5}  & $\setminus$& 2.5                                \\\cline{3-12}
\multicolumn{1}{c|}{}                     &                     &\cellcolor[HTML]{D1DED0}PoseGait\cite{liao2020model}     &\cellcolor[HTML]{D1DED0}0.0  &\cellcolor[HTML]{D1DED0} $\setminus$  &\cellcolor[HTML]{D1DED0}0.2  &\cellcolor[HTML]{D1DED0} $\setminus$ &\cellcolor[HTML]{D1DED0}0.0 & \cellcolor[HTML]{D1DED0}$\setminus$ &\cellcolor[HTML]{D1DED0}0.0 & \cellcolor[HTML]{D1DED0}$\setminus$ & \cellcolor[HTML]{D1DED0}0.1                                        \\
\multicolumn{1}{c|}{}                     &                     &\cellcolor[HTML]{D1DED0}JointsGait\cite{Li2020}          &\cellcolor[HTML]{D1DED0}1.8  &\cellcolor[HTML]{D1DED0}$\setminus$  &\cellcolor[HTML]{D1DED0}2.2   & \cellcolor[HTML]{D1DED0}$\setminus$ &\cellcolor[HTML]{D1DED0}3.5 &\cellcolor[HTML]{D1DED0}$\setminus$ &\cellcolor[HTML]{D1DED0}1.9  &\cellcolor[HTML]{D1DED0}$\setminus$ &\cellcolor[HTML]{D1DED0}2.4                                      \\
\multicolumn{1}{c|}{}                     &                    &\cellcolor[HTML]{D1DED0}SkeletonGait     &\cellcolor[HTML]{D1DED0}\textbf{6.1}   & \cellcolor[HTML]{D1DED0}$\setminus$  & \cellcolor[HTML]{D1DED0}\textbf{{\color[HTML]{FD6864}7.1}}  & \cellcolor[HTML]{D1DED0}$\setminus$ &\cellcolor[HTML]{D1DED0}\textbf{6.3}   &\cellcolor[HTML]{D1DED0}$\setminus$ &\cellcolor[HTML]{D1DED0}\textbf{{\color[HTML]{FD6864}4.6}} & \cellcolor[HTML]{D1DED0}$\setminus$ &\cellcolor[HTML]{D1DED0}\textbf{{\color[HTML]{FD6864}6.0}}  \\ \cline{1-12}
\hline\hline

\end{tabular}
}
\centering
\label{table2}
\end{table}

As shown in Table \ref{table2}, SkeletonGait outperformed advanced model-based methods Posegait\cite{liao2020model} and JointsGait \cite{Li2020} under all occlusion conditions. In BOB, SOB, and MCL, relatively low-level occlusion for human legs, the model-based methods were inferior to the appearance-based methods because the appearance-based methods use high dimension silhouettes as input. With the increase of occlusion degree, the accuracy of all models decreased, but the appearance-based methods decreased the most. In LOB, HCL, and MG with severe occlusion, SkeletonGait surpassed all advanced appearance-based methods. Because appearance-based methods rely heavily on silhouettes, and occlusion always seriously affects silhouettes, while SkeletonGait can deal with occlusion effectively using gait features from the human dual skeleton model.
 
Among all views, the self-occlusion of 90\degree and 270\degree are the most serious. Hence, taking 90\degree as an example, the accuracy of MT3D is 79.8\% (SkeletonGait 43.8\%) in BOB with lowest level of occlusion, which was the highest among all models. However, it decreased to 32.2\% (SkeletonGait 38.2\%) in LOB, 25.9\% (SkeletonGait 27.8\%) in HCL, and even 11.7\% (SkeletonGait 35.8\%) in S of MG. That is to say, with the increase of occlusion degree, the accuracy of MT3D decreased from 79.8\% to 11.7\%, with a decrease rate of more than 85\%, while the accuracy of SkeletonGait decreased from 43.8\% to 27.8\%, with a decrease rate of 36.5\%, indicating that SkeletonGait is more robust to occlusion. Moreover, in the most challenging multi-gait recognition, SkeletonGait had the best performance. In S, its accuracy was 35.8\%, 24.1\% higher than MT3D (11.7\%). However, the accuracy of all appearance-based methods were close to 0\% at 90\degree in T of MG, because when three people walk together at 90\degree, each other's passive occlusion is particularly serious. Even if these silhouettes are obtained from advanced pose-based human instance segmentation method Pose2Seg \cite{2019Pose2Seg} for multi-gait, they are still greatly affected by occlusion, while the accuracy rate of SkeletonGait was 7.1\% under the 90\degree in T of MG. Although it is still very low, it brings hope for multi-gait recognition under occlusion.

Besides, we compare the complexity of these models as shown in Table \ref{table3}. The model parameters of SkeletonGait are the smallest compared with state-of-arts, about 0.52 M, which are about one ninth of appearance-based method MT3D\cite {lin2020gait} (about 4.3 M) and even are about one percent of Posegait \cite{liao2020model} (about 5.04 M). Moreover, SkeletonGait has the fastest convergence speed, only needs about 20K iterations, so the total training time of SkeletonGait still is the shortest. In a word, as the baseline of OG RGB+D database, SkeletonGait has low complexity and is relatively robust to occlusion compared with state-of-art gait recognition methods.

\begin{table}[htb]
\caption{Comparison of complexity between state-of-arts and SkeletonGait.}
\begin{tabular}{c|c|c|c}
\hline\hline
Method       & \begin{tabular}[c]{@{}c@{}}Params (M)\end{tabular} & \begin{tabular}[c]{@{}c@{}}TrainingTime (s/liter)\end{tabular} & Iterations \\ \hline
Gaitset\cite{chao2019gaitset}      & 2.56                                                       & 0.57                                                                   & 80K        \\ 
GaitPart\cite{fan2020gaitpart}     & 1.47                                                       & 0.72                                                                   & 80K        \\ 
MT3D \cite{lin2020gait}         & 4.30                                                       & 1.93                                                                   & 60K        \\ \hline
Posegait \cite{liao2020model}    & 50.4                                                       &\textbf{0.10}                                                                    & 50K        \\ 
JointsGait\cite{Li2020}        & \textbf{0.45}                                                     & 0.66                                                                   & 80K        \\ 
SkeletonGait                   & 0.52                                                     & 0.51                                                                   & \textbf{20K}          \\ \hline \hline
\end{tabular}
\label{table3}
\end{table}

\subsection{Evaluation results on most popular CASIA-B database}

We report the performance of state-of-arts and SkeletonGait under two occlusion-related walking conditions BG and CL of CASIA-B database in Table \ref{table4}. In BG, the subjects carry bags, which is similar to in the BOB and SOB of OG RGB+D database. In CL, the subjects wear coats, which is similar to in the MCL of OG RGB+D database. Therefore, occlusions under BG and CL are not very serious, especially for human legs, so these occlusions have relatively little effect on appearance, which is more conducive to the application of appearance-based methods.

\begin{table}[htb]
\caption{Comparison results of state-of-arts and SkeletonGait on public gait database CASIA-B. In each view, the highest accuracy of each type of methods is reported in \textbf{bold} and highest accuracy of all methods is  reported in \textbf{\textcolor[RGB]{253,104,100}{orange}}.}
\centering
\begin{threeparttable} 
\begin{tabular}{c|c|ccc|c}
\hline\hline
\multicolumn{2}{c|}{}                                                                                                       & \multicolumn{4}{c}{Accuracy(\%)} \\ \cline{3-6} 
\multicolumn{2}{c|}{Gallery}                                                                                                & \multicolumn{4}{c}{0°-180°}      \\ \hline
\multicolumn{2}{c|}{Probe}                                                                                                  & 0°     & 90°   & 180°  & Average  \\ \hline
\multirow{12}{*}{\begin{tabular}[c|]{@{}c@{}}BG\\    \\ \#1-2\end{tabular}} & SPAE\cite{SPAE}             & 34.3   & 40.0  & 32.6  & 35.6     \\
                                                                           & Gaitganv1\cite{GaitGAN}       & 28.6   & 33.5  & 23.6  & 28.6     \\
                                                                           & MGAN\cite{MGAN}               & 48.5   & 49.8   & 43.1  & 47.1     \\
                                                                           & Gaitganv2\cite{Gaitganv2}     & 33.1   & 36.3  & 28.5  & 32.6     \\
                                                                           & Gaitset\cite{chao2019gaitset} & 79.9   & 76.7  & 73.0  & 76.5     \\
                                                                           & MT3D\cite{lin2020gait}        & \textbf{{\color[HTML]{FD6864} 86.7}}    & \textbf{{\color[HTML]{FD6864}82.5}}   & \textbf{{\color[HTML]{FD6864}81.2}} & \textbf{{\color[HTML]{FD6864} 83.5}}    \\ \cline{2-6} 
                                                                           & \cellcolor[HTML]{D1DED0} PTSN\cite{PTSN}               & \cellcolor[HTML]{D1DED0}22.4   &\cellcolor[HTML]{D1DED0}31.5  &\cellcolor[HTML]{D1DED0}18.2  & \cellcolor[HTML]{D1DED0}24.0     \\
                                                                           & \cellcolor[HTML]{D1DED0}PTSN-3D\cite{PTSN-3D}                             &\cellcolor[HTML]{D1DED0}27.7   & \cellcolor[HTML]{D1DED0}37.5  & \cellcolor[HTML]{D1DED0}27.0  &\cellcolor[HTML]{D1DED0}30.7     \\
                                                                           & \cellcolor[HTML]{D1DED0}PoseGait\cite{liao2020model}  & \cellcolor[HTML]{D1DED0}29.1   & \cellcolor[HTML]{D1DED0}42.2  & \cellcolor[HTML]{D1DED0}26.7  & \cellcolor[HTML]{D1DED0}32.7     \\
                                                                           & \cellcolor[HTML]{D1DED0}JointsGait\cite{Li2020}       & \cellcolor[HTML]{D1DED0}54.3   & \cellcolor[HTML]{D1DED0}65.7  & \cellcolor[HTML]{D1DED0}50.1  & \cellcolor[HTML]{D1DED0}56.7     \\
                                                                           & \cellcolor[HTML]{D1DED0}SkeletonGait\tnote{1}                                   & \cellcolor[HTML]{D1DED0}68.7   &\cellcolor[HTML]{D1DED0}71.0    & \cellcolor[HTML]{D1DED0}69.5  & \cellcolor[HTML]{D1DED0}69.7     \\
                                                                           & \cellcolor[HTML]{D1DED0}SkeletonGait\tnote{2}                                 & \cellcolor[HTML]{D1DED0}\textbf{70.1}   & \cellcolor[HTML]{D1DED0}\textbf{73.3} & \cellcolor[HTML]{D1DED0}\textbf{71.1} & \cellcolor[HTML]{D1DED0}\textbf{71.5}     \\ \hline
\multirow{12}{*}{\begin{tabular}[c|]{@{}c@{}}CL\\    \\ \#1-2\end{tabular}} & SPAE\cite{SPAE}             & 21.5   & 22.2  & 19.6  & 21.1     \\
                                                                           & Gaitganv1\cite{GaitGAN}       &9.8    & 19.9  & 11.9  & 13.9     \\
                                                                           & MGAN\cite{MGAN}               & 23.1   & 32.7  & 21.0  & 25.6     \\
                                                                           & Gaitganv2\cite{Gaitganv2}     & 11.3   & 22.5  & 13.3  & 15.7     \\
                                                                           & Gaitset\cite{chao2019gaitset} & 52.0   & 61.2  & 45.9  & 53.0     \\
                                                                           & MT3D\cite{lin2020gait}        & \textbf{{\color[HTML]{FD6864}67.5}}   &\textbf{{\color[HTML]{FD6864} 69.8}}    & \textbf{59.0}  & \textbf{{\color[HTML]{FD6864} 65.4}}    \\ \cline{2-6} 
                                                                           & \cellcolor[HTML]{D1DED0}PTSN\cite{PTSN}               & \cellcolor[HTML]{D1DED0}14.2   & \cellcolor[HTML]{D1DED0}20.0  & \cellcolor[HTML]{D1DED0}14.0  & 1\cellcolor[HTML]{D1DED0}6.1     \\
                                                                           &\cellcolor[HTML]{D1DED0}PTSN-3D\cite{PTSN-3D}                             &\cellcolor[HTML]{D1DED0}15.8   & \cellcolor[HTML]{D1DED0}24.3  & \cellcolor[HTML]{D1DED0}17.0  & \cellcolor[HTML]{D1DED0}19.0     \\
                                                                           & \cellcolor[HTML]{D1DED0}PoseGait\cite{liao2020model}  & \cellcolor[HTML]{D1DED0}21.3   & \cellcolor[HTML]{D1DED0}34.9  & \cellcolor[HTML]{D1DED0}19.7  & \cellcolor[HTML]{D1DED0}25.3     \\
                                                                           & \cellcolor[HTML]{D1DED0}JointsGait\cite{Li2020}       & \cellcolor[HTML]{D1DED0}48.1   & \cellcolor[HTML]{D1DED0}52.3  & \cellcolor[HTML]{D1DED0}52.0  & \cellcolor[HTML]{D1DED0}50.8     \\
                                                                           & \cellcolor[HTML]{D1DED0}SkeletonGait\tnote{1}                                   &\cellcolor[HTML]{D1DED0}62.5   & \cellcolor[HTML]{D1DED0}67.8  & \cellcolor[HTML]{D1DED0}61.4  & \cellcolor[HTML]{D1DED0}63.9     \\
                                                                           & \cellcolor[HTML]{D1DED0}SkeletonGait\tnote{2}                                 &\cellcolor[HTML]{D1DED0}\textbf{63.1}    & \cellcolor[HTML]{D1DED0}\textbf{ 69.2 } & \cellcolor[HTML]{D1DED0}\textbf{{\color[HTML]{FD6864} 62.4}}   & \cellcolor[HTML]{D1DED0}\textbf{64.9}     \\ \hline\hline

\end{tabular}
\begin{tablenotes}    
        \footnotesize              
        \item[1] SkeletonGait uses the joints  directly estimated by the pose estimation model of work\cite{moon2019camera}.          
        \item[2] SkeletonGait uses the joints estimated by the pose estimation model of work\cite{moon2019camera} that is trained again on OG RGB+D database.        
      \end{tablenotes}            
    \end{threeparttable}      
\centering
\label{table4}
\end{table}
However, as shown in Table \ref {table4}, SkeletonGait, trained on 62 subjects of CASIA-B database according to other models, was far superior to other model-based gait recognition methods and better than most appearance-based methods. Even was close to appearance-based MT3D \cite {lin2020gait} in CL which achieved the highest accuracy. However, when the occlusion is serious, for example, in LOB, HCL and MG of OG RGB+D databse as shown in Table \ref{table2}, the recognition accuracy of the appearance-based methods were seriously reduced, which not as robust to occlusion as SkeletonGait. Because silhouettes that the appearance-based methods rely on are easily affected by the environment, such as shadow and occlusion, while joints are easier to obtain than silhouettes in real life with the help of human pose estimation methods or Kinect. Besides, it can be seen from the experimental results that the post-training estimation model of work \cite {moon2019camera} trained by the OJ data of OG RGB+D can more accurately represent the human body, and the average gait recognition accuracy is improved by 1.4\% in BG and CL. In a word, the model-based SkeletonGait not only has satisfactory performance, but also is easier to apply in practice.

\subsection{ Comparison between CASIA-B database and OG RGB+D database}

\begin{figure}[htb]
\begin{center}

\includegraphics[width=0.6 \linewidth]{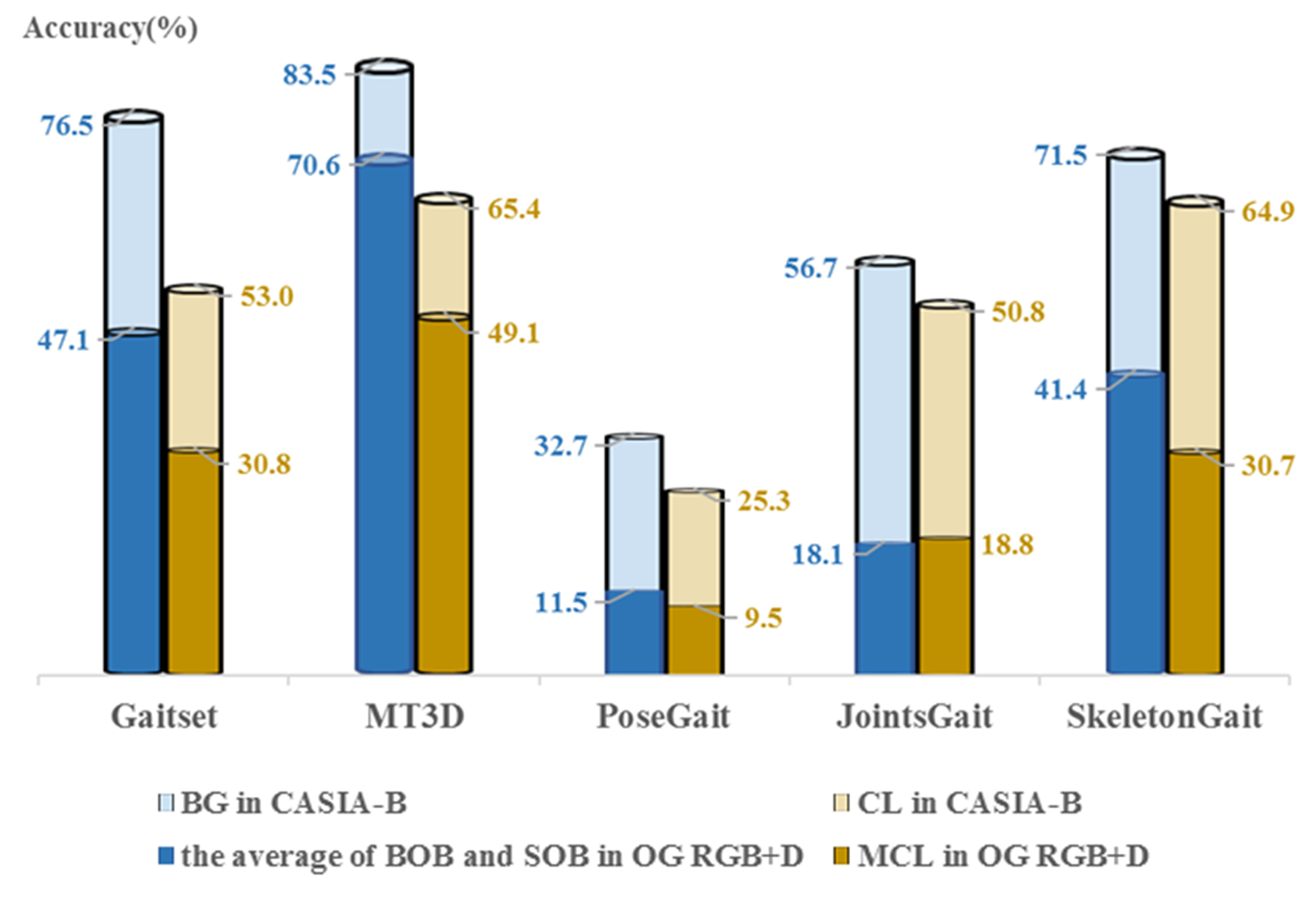}

\end{center}
\caption{The performance comparisons of gait recognition methods on OG RGB+D database and CASIA-B database. }
\label{fig11}
\end{figure}

Because the BG of CASIA-B database is similar to the BOB and SOB of OG RGB+D database, and the CL of CASIA-B database is similar to the MCL of OG RGB+D database, we compare the performance of appearance-based gait recognition methods Gaitset\cite{chao2019gaitset}, MT3D\cite{lin2020gait} and model-based methods PoseGait\cite{liao2020model}, JointsGait\cite{Li2020}, SkeletonGait on OG RGB+D database and CASIA-B database as shown in Fig.\ref{fig11}. It can be seen from Fig.\ref{fig11} that although the occlusion degree of the two databases is similar, the accuracy of all methods in OG RGB+D database were significantly lower than that in CASIA-B database. Because video clips range from 12 frames to 80 frames in OG RGB+D database, the number of irregular samples increases the difficulty of gait recognition.

\begin{figure}
\begin{center}

\includegraphics[width=0.8 \linewidth]{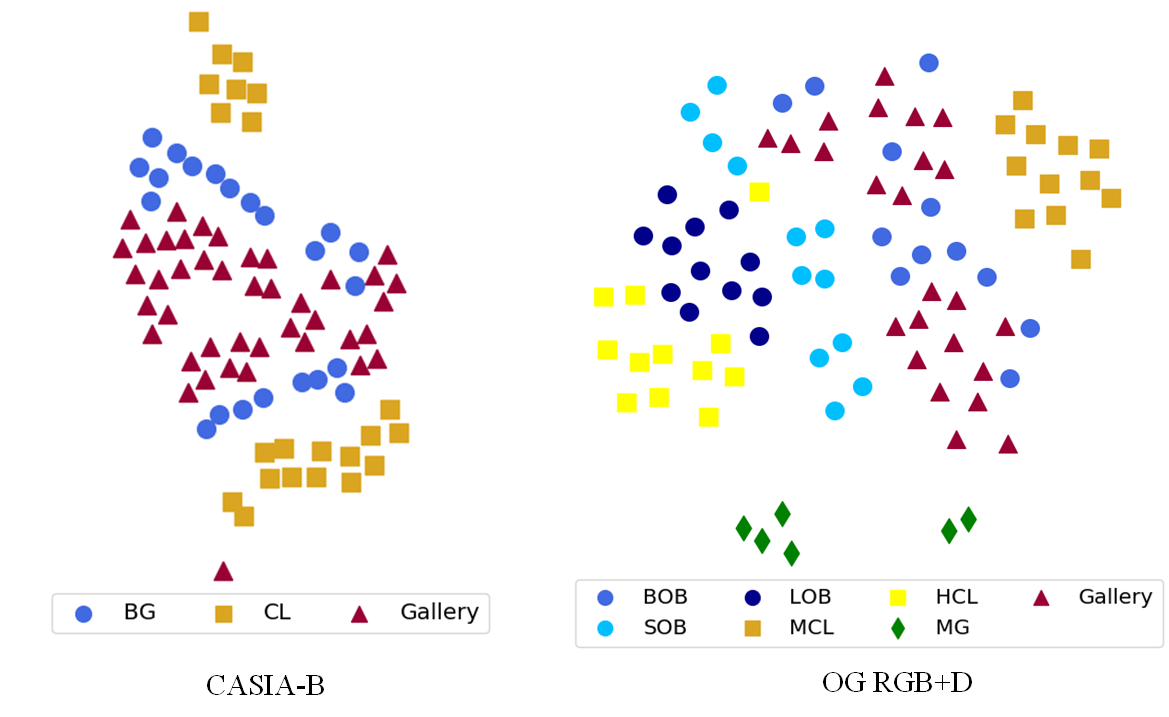}

\end{center}
\caption{The gait feature distributions comparisons of CASIA-B database and OG RGB+D database.}
\label{fig12}
\end{figure}

Moreover, as shown in Fig.\ref{fig12}, we compared the gait feature distributions of the walking conditions contained in the two databases. It is obvious that the feature distributions of BG and CL in CASIA-B database are very close to the feature distribution of gallery, while the occlusion conditions contained in OG RGB+D database are not only more than CASIA-B database, but also their feature distributions are far away from the feature distribution of gallery, especially LOB, HCL and MG. Therefore, OG RGB+D database is more challenging for gait recognition, which is helpful to the research of occlusion robust gait recognition methods.

\section{ Conclusion }
In this paper, we introduce a new benchmark OG RGB+D database, which breaks through the limitation of other gait databases and focuses on gait recognition under occlusion. This database contains common occlusions in our daily life, including self-occlusion (eight views), active occlusion (backpack, carried small belongings, carried large belongings, heavy coat, long coat or long skirts), and passive occlusion (three subjects walking together) ignored by other gait recognition databases. To collect OG RGB+D database, we design a data acquisition system multi-Kinect SDAS that is composed of three parts, including a hardware system based on three synchronous Azure Kinect DK sensors, an acquisition software and a multi-view data fusion processing method, and this system can also be applied to all occasions where the human pose needs to be accurately estimated. Thanks to multi-Kinect SDAS, OG RGB+D database contains multi-modal gait data under occlusion, especially contains the OJ data which can more accurately represent the human body and silhouettes obtained by using advanced technology, which can  promote to develop different types of robust gait recognition methods.

Besides, we propose a novel model-based gait recognition method called SkeletonGait, which learns more discriminative gait information from the human dual skeleton model based on siamese-STGCN. The experimental results demonstrate that SkeletonGait greatly outperforms advanced model-based methods in all occlusion conditions, and performs superior to state-of-art appearance-based methods when occlusions seriously affect people’s appearance. Besides,  Our model has small parameters, fast convergence speed, and good stability. Hence, effective gait features extracted from human body structure and movements have great potential for robust gait recognition, especially for multi-gait. In the future, we will continue to work on gait recognition under occlusion and strive to improve its accuracy.






\end{document}